\documentclass[10pt,twocolumn,letterpaper]{article}

\usepackage[pagenumbers]{cvpr} 

\input{preamble}
\usepackage{colortbl}
\usepackage{array}
\definecolor{scbase}{HTML}{65758A}
\definecolor{scft}{HTML}{D9711A}
\definecolor{scrl}{HTML}{3A6CD4}
\definecolor{scteal}{HTML}{0E9384}
\definecolor{scred}{HTML}{D92D20}
\definecolor{oursrow}{HTML}{E6F4F1}
\definecolor{panelline}{HTML}{D0D5DD}
\definecolor{mutedtext}{HTML}{667085}
\newcommand{\ctitle}[1]{\textbf{#1}}
\newcommand{\ours}{\rowcolor{oursrow}}
\newcommand{\panelhead}[1]{\textit{#1}}
\newcolumntype{C}{>{\centering\arraybackslash}X}
\newcolumntype{L}{>{\raggedright\arraybackslash}X}
\newlength{\pw}
\newcommand{\gpanel}[2][panelline]{\raisebox{-.5\height}{\setlength{\fboxsep}{0pt}\setlength{\fboxrule}{0.7pt}%
  \fcolorbox{#1}{white}{\includegraphics[width=\dimexpr\pw-1.4pt\relax]{#2}}}}
\newcommand{\gpanelc}[2]{\raisebox{-.5\height}{\setlength{\fboxsep}{0pt}\setlength{\fboxrule}{1.3pt}%
  \fcolorbox{#1}{white}{\includegraphics[width=\dimexpr\pw-2.6pt\relax]{#2}}}}
\newcommand{\pill}[2]{{\setlength{\fboxsep}{2pt}\colorbox{#1}{\textcolor{white}{\sffamily\bfseries\fontsize{8.5}{9.5}\selectfont\vphantom{Tg}\,#2\,}}}}
\newcommand{\glabel}[1]{{\sffamily\fontsize{9}{10.5}\selectfont #1}}
\newcommand{\gnote}[1]{{\sffamily\fontsize{9}{10.5}\selectfont\color{mutedtext}#1}}
\newcommand{\quotestyle}{\centering\sffamily\itshape\fontsize{9}{10.5}\selectfont\color{mutedtext}}
\newcommand{\vlabel}[1]{\rotatebox[origin=c]{90}{#1}}

\definecolor{cvprblue}{rgb}{0.21,0.49,0.74}
\usepackage[pagebackref,breaklinks,colorlinks,allcolors=cvprblue]{hyperref}
\def\confName{CVPR}
\def\confYear{2027}

\title{\papertitle}

\author{
Zhenghao Ni$^{1}$ \qquad Weimin Qiu$^{2}$ \qquad Justin Ngo$^{2}$ \qquad Meng Tang$^{2}$\\
$^{1}$University of Toronto \qquad $^{2}$University of California, Merced\\
{\tt\small zhenghao.ni@mail.utoronto.ca, \{wqiu5, jngo50, mtang4\}@ucmerced.edu}\\
{\small Project page: \url{https://sc-video-reasoning.github.io}}
}

\begin{document}
\twocolumn[{%
\renewcommand\twocolumn[1][]{#1}%
\maketitle
\begin{center}
\centering
\includegraphics[width=\linewidth]{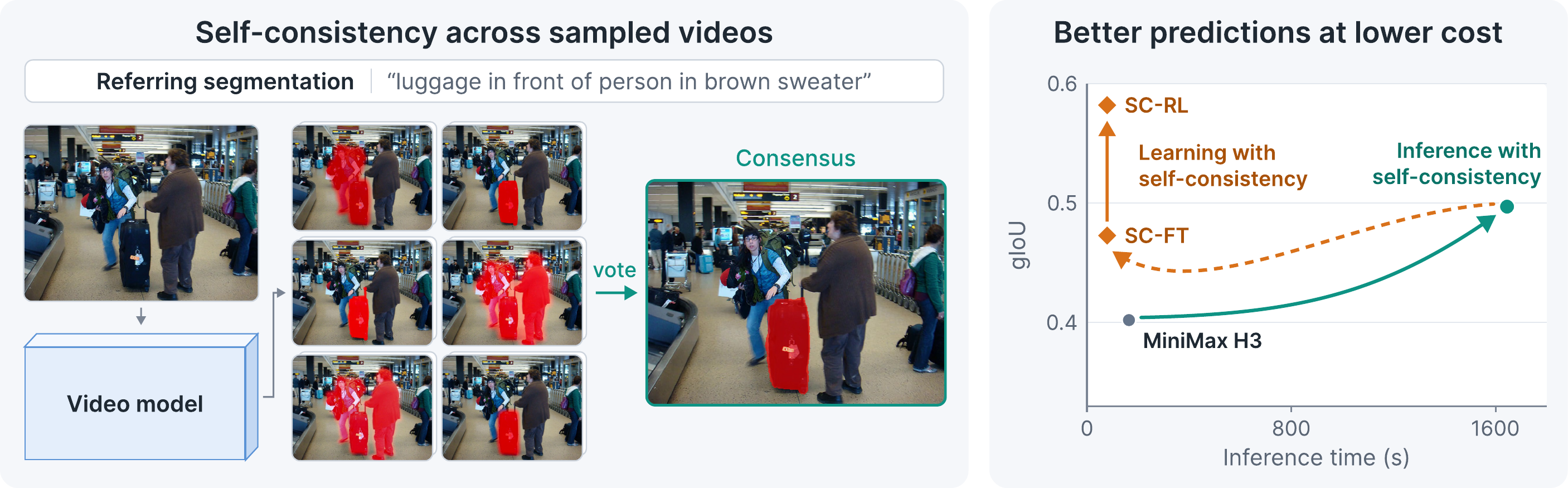}
\vspace{-2pt}
\captionsetuptype
\captionsetup{hypcap=false}%
\caption{\meng{Although self-consistency improves video reasoning at inference time, generating multiple rollouts substantially increases latency. We use consensus predictions to post-train the model, giving better predictions at reduced latency.}
(Left) Videos sampled for one prompt disagree: some mask only the referred luggage, others also mask the people next to it, while their consensus keeps the luggage.
(Right) Inference with self-consistency improves MiniMax-H3 predictions through aggregation; learning with self-consistency (SC-FT, then SC-RL) turns consensus into faster and, with SC-RL, stronger single-generation predictions.
}
\label{fig:teaser}
\end{center}
\vspace{2pt}

}]
\begin{abstract}
While video diffusion models have demonstrated emerging zero-shot capability for visual reasoning and perception tasks,
their stochastic generations often produce inconsistent predictions for tasks requiring reliable, structured outputs.
\meng{Prior work has shown that self-consistency across stochastic video generations can improve diffusion-based reasoning at test time.}
\meng{We systematically extend the test-time method to structured prediction problems such as referring segmentation.}
\meng{However, test-time scaling via self-consistency requires multiple rollouts and incurs much more latency.}
We propose self-consistency as an intrinsic learning signal for post-training video diffusion models without ground-truth annotations or external verifiers.
Specifically, \meng{our proposed} Self-Consistency Fine-Tuning (SC-FT) distills consensus predictions into the video generation model, while subsequent Self-Consistency Reinforcement Learning (SC-RL) rewards generations according to their agreement with the consensus.
The resulting model effectively internalizes the benefit of multi-sample consensus and requires only a single generation at inference time, while substantially outperforming the original model.
To \meng{further} reduce inference overhead, we read out predictions early in the denoising trajectory, preserving consensus quality while cutting inference time per rollout by 52\%.
For visual search, \meng{our method} raises task accuracy from 48.4\% for a single generation to \textbf{99.1\%}.
For maze solving across three layout sizes, SC-FT and SC-RL \meng{improve} single-generation strict validity from 59.1\% to 66.1\% and \textbf{95.6\%} with the same inference latency.
Also, for referring segmentation, SC-RL raises single-generation gIoU from 0.402 to \textbf{0.582}, beyond SC-FT's 0.473.

\end{abstract}

\section{Introduction}
\label{sec:introduction}
Video generation models are emerging general-purpose models for computer vision tasks such as perception, reasoning, and physics modeling.
Recent work demonstrates video generators as zero-shot learners and reasoners~\citep{veo3} and successful adaptation to various vision tasks~\citep{wang2026genception}.
An image-text-to-video model provides a common interface for many tasks, e.g., referring expression segmentation and maze solving.
However, an extensive empirical study~\citep{guo2026video} reveals that even leading video generators such as Veo~3~\citep{deepmind2025veo} are not yet reliable as zero-shot reasoners.

\meng{Prior work~\cite{tong2025thinkingwithvideo,wang2026demystifying} has improved video reasoning by aggregating predictions across stochastic generations at inference time, \textit{à la} \textbf{self-consistency} in chain-of-thought reasoning~\citep{llmselfconsistency}.}
While diversity is desirable in stochastic video generation for content creation, variability in task-relevant predictions can undermine the reliability required by structured visual reasoning and perception tasks, which often have a unique solution.
\meng{Figure~\ref{fig:teaser} illustrates that consensus among multiple rollouts improves inference for referring expression segmentation.}
\meng{We extend such test-time scaling methods to video reasoning with structured predictions} \rebuttal{such as spatial locations, trajectories, and segmentation masks. We measure agreement in task-specific output spaces rather than directly comparing generated videos (Figure~\ref{fig:method-pipeline})}.

\meng{A key limitation of test-time scaling is increased inference latency, as multiple rollouts must be generated.}
\rebuttal{We therefore seek to internalize the benefits of multi-sample consensus into video generation models through \textit{post-training}, enabling reliable reasoning with only a single rollout at inference time.}
\meng{Previous post-training methods} for video reasoning through generation include supervised fine-tuning~\citep{yang2025vrbench,wang2026genception} and reinforcement learning with verifiable rewards (RLVR)~\citep{zhu2026video}.
However, these approaches typically require task-specific ground-truth supervision or external verifiers.

\rebuttal{We investigate whether agreement across stochastic video generations can provide a reliable training signal for video reasoning without requiring ground-truth supervision or external verification.}
Specifically, we first propose Self-Consistency Fine-Tuning (SC-FT), which fine-tunes a pretrained video generator on its own consensus predictions, in the spirit of Rejection Fine-Tuning~\citep{yuan2023scaling}.
Our method requires neither ground-truth annotations nor external verification, making it naturally suited for scalable self-improvement.
To reduce confirmation bias~\citep{arazo2020pseudo}, a common failure mode of self-training, we retain only pseudo-solutions with sufficiently high cross-generation agreement for fine-tuning.
In this way, our method distills the consensus obtained from multiple stochastic rollouts into a single model, enabling efficient inference with only one rollout.

\rebuttal{Beyond SC-FT, we also introduce Self-Consistency Reinforcement Learning (SC-RL), which further improves the student by rewarding its own stochastic generations according to their agreement with consensus targets.
We \meng{adapt} DiffusionART~\citep{diffusionart} to optimize the video diffusion model using these intrinsic rewards (Sec.~\ref{sec:sc_art}).}

\rebuttal{We evaluate \meng{our post-training methods} on three tasks with distinct output structures: visual search (spatial locations), maze solving (trajectories), and referring segmentation (pixel-level masks).}
\meng{To further reduce latency, we read out each prediction early in the denoising trajectory, building on evidence that video models commit to their answers early~\citep{newman2026videomodelsreasonearly,wang2026demystifying}.}
Extensive experiments demonstrate substantial improvement with our post-training methods.

Our sample--aggregate--filter--distill recipe is simple and improves all three tasks.
SC-FT improves single-generation segmentation gIoU from 0.402 to \textbf{0.473}, search accuracy from 48.4\% to \textbf{78.2\%}, and maze strict validity from 59.1\% to \textbf{66.1\%}.
\rebuttal{SC-RL further raises segmentation gIoU to \textbf{0.582}, maze strict validity to \textbf{95.6\%}, and visual search accuracy to \textbf{99.1\%}.}
\revorange{It also generalizes to unseen $7\times7$ and $8\times8$ mazes (Figure~\ref{fig:maze-qualitative}).}
Our main contributions are summarized as follows:
\begin{itemize}[leftmargin=10pt]
\item We identify and systematically characterize self-consistency as an intrinsic signal for video reasoning, and extend test-time methods to structured, continuous outputs (segmentation, marker locations, and trajectories).%
\item \meng{Our primary contribution is post-training methods for diffusion-based video reasoning, so that a single rollout suffices during inference. Our proposed SC-FT and SC-RL are based on self-consistency of stochastic generation}.
\item Our methods substantially improve performance while reducing latency for multiple tasks including maze solving, visual search, and referring segmentation.
\end{itemize}

\section{Related Work}
\label{sec:related}

\vspace{-3pt}
\paragraph{Video generation for perception and reasoning.}
Video generation expresses solutions to perception and reasoning tasks through generated frame sequences.
GenCeption~\citep{wang2026genception} adapts video generation priors for visual tasks, while InterPose~\citep{cai2025interpose} selects pose estimates through geometric consistency across generated interpolation videos.
Veo~3 demonstrates zero-shot visual reasoning~\citep{veo3}, and VR-Bench~\citep{yang2025vrbench} evaluates spatial planning through maze solving and shows that supervised fine-tuning on solution videos improves this capability.
Wan-R1~\citep{liu2026wanr1} and VideoRLVR~\citep{zhu2026video} further optimize generated trajectories with verifiable rewards for spatial planning and rule-based puzzles.
Our work connects inference and post-training through structured output consensus: agreement among paths, points, and masks from a frozen video generator provides both aggregated predictions and training targets.
Distilling this consensus \meng{through fine-tuning and RL} improves single-generation prediction without ground-truth videos~\citep{wang2026genception} or external verifiers~\citep{zhu2026video}.

\vspace{-3pt}
\paragraph{Self-consistency for inference and self-training.}
Self-consistency~\citep{llmselfconsistency} improves chain-of-thought reasoning for LLMs by sampling diverse reasoning paths and selecting the final answer by majority voting.
This principle also supports self-training: LMSI~\citep{huang2023selfimprove} fine-tunes on reasoning paths that agree with the majority answer, and TTRL~\citep{zuo2026ttrl} turns majority-vote answers into pseudo-rewards for test-time reinforcement learning.
MM-UPT~\citep{wei2025mmupt} extends this reward construction to multimodal LLMs, while SCRL~\citep{yan2026if} refines pseudo-label supervision through consensus-based selection and entropy-gated negative labeling.
\rebuttal{Beyond LLMs, }
\meng{Demystifying Video Reasoning~\cite{wang2026demystifying}} shows Training-Free Ensemble of diffusion latents.
Thinking with Video~\citep{tong2025thinkingwithvideo} observes that majority voting over repeated Sora-2 generations improves accuracy on verifiable puzzles with discrete answers, and identifies test-time scaling for video reasoning as an underexplored direction.
We study this direction systematically and extend to structured, continuous outputs (masks, points, and paths) where exact-match voting does not apply, and reduce its cost through early clean-latent readout.
\meng{To address the key limitation of increased latency, we propose post-training methods by leveraging self-consistency signal.}

\vspace{-3pt}
\revorange{\paragraph{Reward and preference post-training for diffusion models.}
DDPO~\citep{black2024ddpo} optimizes denoising trajectories with policy gradients from a black-box reward function, Diffusion-DPO~\citep{wallace2024diffusiondpo} aligns the denoiser on sample pairs ranked by human preferences or a reward model, and DGPO~\citep{luo2026dgpo} splits sampled rollouts into positive and negative groups by reward score for direct group preference optimization.
All three methods depend on external rewards such as human preference labels, learned reward models, or task verifiers to score generated samples.
In contrast, we obtain training targets and rewards from agreement across the model's own rollouts, without external rewards or ground-truth labels.}
\vspace{-5pt}
\section{Method}
\label{sec:method}
\begin{figure*}[t!]
\centering
\includegraphics[width=\linewidth]{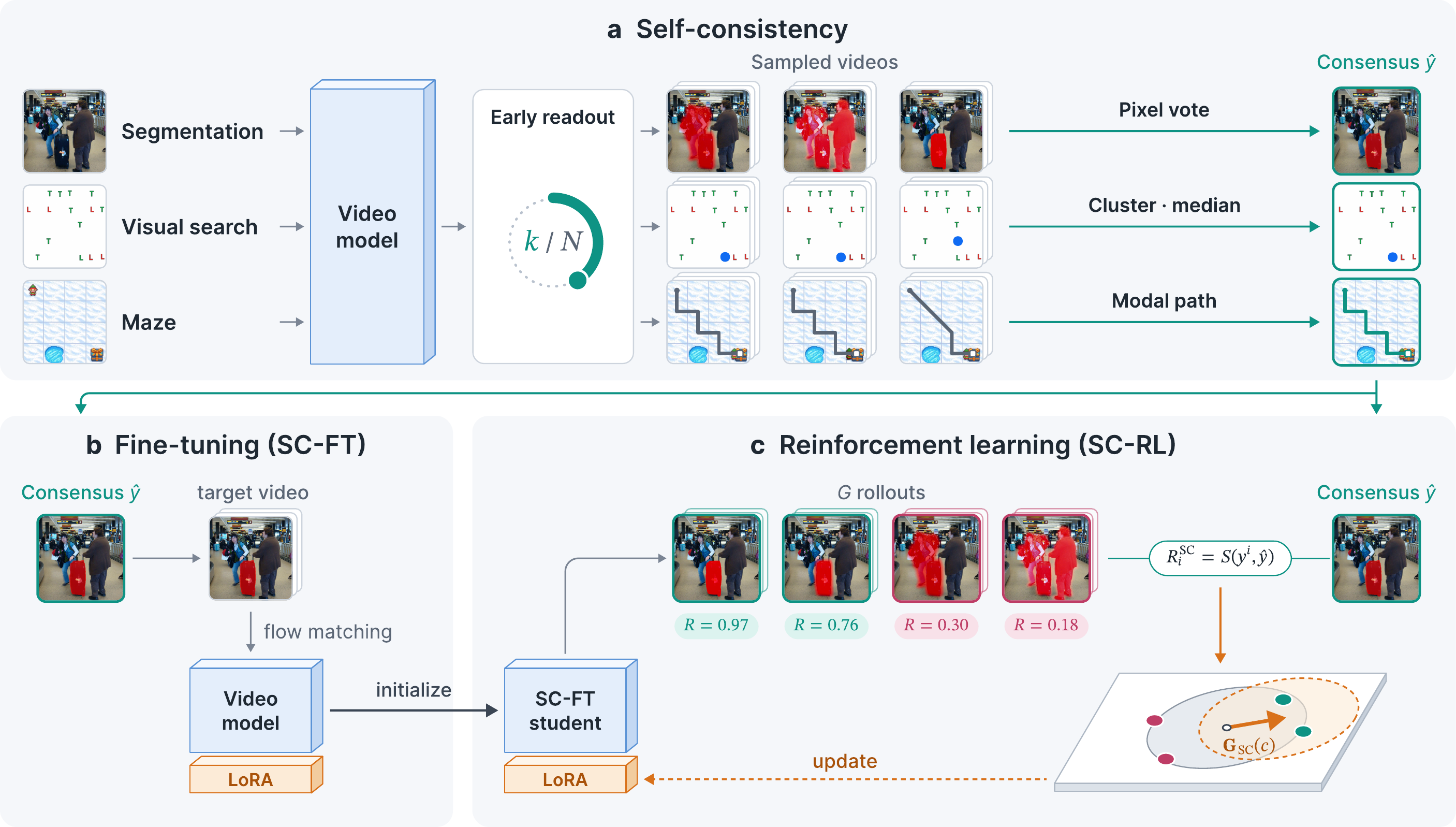}
\caption{Learning via self-consistency.
(a) Test-time scaling with self-consistency: a frozen video model generates $G$ rollouts per input, and the answers extracted from them are aggregated into a consensus $\widehat{y}$.
Early readout decodes at step $k$ of $N$, cutting denoiser calls from $NG$ to $kG$.
(b) SC-FT trains a LoRA student on target videos rendered from the consensus (the unchanged input for abstention); mazes reuse rollouts that follow the consensus path.
(c) SC-RL rewards each rollout of the SC-FT student for agreeing with the consensus, $R_i^{\mathrm{SC}}=S(y^i,\widehat{y})$, and combines the rewards into a direction $\mathbf{G}_{\mathrm{SC}}(c)$ that updates the policy.}
\label{fig:method-pipeline}
\end{figure*}

We propose two post-training methods based on self-consistency, including \textbf{Self-Consistency Fine-Tuning (SC-FT)} and \textbf{Self-Consistency Reinforcement Learning (SC-RL)} detailed in Sec.~\ref{sec:self-consistency-posttraining} and Sec.~\ref{sec:sc_art} respectively.
We introduce the problem setting and reasoning tasks in Sec.~\ref{sec:structured-predictions}, and give task-specific methods for obtaining a consensus solution in Sec.~\ref{sec:multi-seed-voting}.
Sec.~\ref{sec:early-voting} shows how to accelerate sampling via early-denoising readout. Fig.~\ref{fig:method-pipeline} illustrates our pipeline.

\vspace{-5pt}
\subsection{Tasks and Structured Predictions}
\label{sec:structured-predictions}

\revorange{Let $c=(I,q)$ be an input image with a text instruction.
A video generator $F_\theta$ maps $c$ and a seed $s_i$ to a video, and a task-specific extractor $L$ reads the answer from it:
\begin{equation}
 V^i=F_\theta(c;s_i),\qquad y^i=L(V^i,c).
 \label{eq:extraction}
\end{equation}
We study three tasks.
In \emph{referring segmentation}, the model paints the described region in red, and $y^i$ is the mask of the overlay.
In \emph{maze solving}, it moves an agent from start to goal while avoiding holes, and $y^i$ is the sequence of cells visited.
In \emph{visual search}, it marks the object with the specified features or leaves the image unchanged when there is none, and $y^i$ is the marker location or the absence symbol $\bot$.
The same recipe applies whenever the answer appears in the video in a form that a fixed extractor can read.
Datasets and protocols are given in Section~\ref{sec:experiments}, and the extractors in Appendix~\ref{app:inference}.}

\subsection{Consensus in Task Output Space}
\label{sec:multi-seed-voting}

\revorange{We compute the consensus over the $G$ predictions in the output space of the task:
\begin{equation}
 \widehat y=A(y^1,\ldots,y^G;c,m).
 \label{eq:general-consensus}
\end{equation}
\rebuttal{where $A$ is the task specific voting function and $m$ is the threshold(we set m=9 for a group of 10 rollouts).} Voting on extracted answers ignores differences in appearance that do not matter for the task.
Unlike discrete answers by large language models, masks, paths, and points rarely coincide exactly, so each task defines its own agreement as follows.}

\paragraph{Points.}
Markers within a radius $r$ of each other form a group.
If the largest group has at least $m$ members, the consensus is its median; otherwise it is $\bot$.
Spurious markers scatter across rollouts, so disagreement itself indicates that no target is present.

\paragraph{Paths.}
Two paths agree when they visit the same cells in the same order.
Inference returns the most frequent path; training keeps it only when it is unique and at least $m$ rollouts follow it.

\paragraph{Masks.}
Agreement is counted per pixel $u$: with $M^i\in\{0,1\}^{H\times W}$ the mask of rollout $i$,
\begin{equation}
\widehat M(u)= \mathbb{I}\Big[\textstyle\sum_{i=1}^{G}M^i(u)\geq m\Big].
\label{eq:pseudo_mask}
\end{equation}
A high $m$ removes regions that only some rollouts paint, which raises precision, and the result is a new mask rather than one of the rollouts.

\revorange{Agreement serves as a proxy for correctness: correct answers recur across rollouts while errors scatter (Section~\ref{sec:ablations}).
$G$, $m$, and $r$ are fixed per task (Appendix~\ref{app:training}).}

\subsection{Early-Denoising Readout}
\label{sec:early-voting}

\revorange{Consensus over $G$ rollouts costs $G$ times the denoising work of one.
The answer, however, depends only on spatial layout: where a marker lands, which cells a path crosses, which region a mask covers.
Such layouts form early in denoising~\citep{newman2026videomodelsreasonearly}, while later steps refine appearance.
We therefore stop the sampler at step $k$ of $N$, decode the predicted clean latent $\widehat z_{\mathrm{clean},k}^i$ with the video decoder $D$, and apply the same extractor and aggregation:
\begin{equation}
 y_k^i=L\!\left(D(\widehat z_{\mathrm{clean},k}^i),c\right),
 \qquad \widehat y_k=A(y_k^1,\ldots,y_k^G;c,m).
 \label{eq:early-consensus}
\end{equation}
This cuts denoiser calls from $NG$ to $kG$ with the seeds and the consensus rule unchanged.
Decoding adds a fixed cost per rollout, so Section~\ref{sec:early-results} reports measured time as well as step counts (Appendix~\ref{app:inference} tests decoding only some rollouts).}

\subsection{Self-Consistency Fine-Tuning (SC-FT)}
\label{sec:self-consistency-posttraining}

\revorange{SC-FT distills the consensus into a student that answers with a single rollout.
A frozen teacher $F_{\theta_0}$ labels unlabeled inputs with its consensus, and a LoRA~\citep{hu2022lora} student $F_\theta$ is trained on the inputs whose consensus passes the support threshold: rejection fine-tuning with agreement in place of a verifier (Algorithm~\ref{alg:consensus-training}, stage 1).}

\begin{algorithm}[t]
\caption{Learning via self-consistency}
\label{alg:consensus-training}
\begin{algorithmic}[1]
\Require Unlabeled inputs $\mathcal C$, frozen model $F_{\theta_0}$
\Statex \hspace{-\algorithmicindent}{\small\textbf{Stage 1: Self-Consistency Fine-Tuning (SC-FT)}}
\For{each $c\in\mathcal C$}
    \State Sample $G$ rollouts from $F_{\theta_0}$ and extract $y^{1:G}$
    \State $\widehat y\gets A(y^1,\ldots,y^G;c,m)$
    \State If the support of $\widehat y$ reaches $m$, add its target $z^\star$ to $\mathcal B$
\EndFor
\State Train a LoRA student $F_\theta$ on $\mathcal B$ with $\mathcal L_{\mathrm{SC}}$ (Eq.~\ref{eq:asc_pl_loss})
\Statex \hspace{-\algorithmicindent}{\small\textbf{Stage 2: Self-Consistency Reinforcement Learning (SC-RL)}}
\State Compute and fix the consensus $\widehat y$ for each $c\in\mathcal C$
\For{each update, with $\theta_{\mathrm{old}}\gets\theta$}
    \State Sample $G$ rollouts per input from $F_{\theta_{\mathrm{old}}}$
    \State Reward each rollout: $R_i^{\mathrm{SC}}\gets S(y^i,\widehat y)$
    \State Compute the direction $\mathbf{G}_{\mathrm{SC}}(c)$ (Eq.~\ref{eq:sc_art_guidance})
    \State Regress $v_\theta$ onto $v_{\theta_{\mathrm{old}}}+\mathbf{G}_{\mathrm{SC}}(c)$
\EndFor
\State \Return $F_\theta$, which answers with one rollout
\end{algorithmic}
\end{algorithm}

\revorange{\paragraph{Targets.}
Each accepted consensus becomes a target video, encoded by the VAE into a clean latent $z^\star$.
Segmentation renders the consensus mask as a red overlay, keeping masks whose foreground fraction lies in a fixed range.
Search draws a blue marker at the consensus point, and an abstention gives the unchanged input as target.
Mazes reuse up to three teacher rollouts that follow the consensus path and take their predicted clean latents as $z^\star$.}

\revorange{\paragraph{Objective.}
With noise at $t=0$ and data at $t=1$~\citep{flowmatching,rectifiedflow}, we set $z_t=t z^\star+(1-t)\epsilon$ for Gaussian $\epsilon$ and regress the student's velocity onto $u_t^\star=z^\star-\epsilon$:
\begin{equation}
 \mathcal L_{\mathrm{SC}}=
 \mathbb E_{c,t,\epsilon}
 \left[\left\|v_\theta(z_t,t,c)-u_t^\star\right\|_2^2\right].
 \label{eq:asc_pl_loss}
\end{equation}
Only the LoRA parameters are updated.
Part of the search training starts from the teacher's stored state $z_t$ instead of fresh noise, with $u_t^\star=(z^\star-z_t)/(1-t)$ (Appendix~\ref{app:training}).}

\subsection{Self-Consistency Reinforcement Learning (SC-RL)}
\label{sec:sc_art}

\rebuttal{Unlike SC-FT, which learns from a fixed set of consensus-derived targets, SC-RL further optimizes the student by rewarding its own rollouts based on their agreement with the consensus. This enables learning beyond the teacher-generated samples. Specifically, SC-RL initializes from the SC-FT student and performs a second stage of post-training using DiffusionART~\cite{diffusionart} with a new LoRA adapter (Figure~\ref{fig:method-pipeline}c; Algorithm~\ref{alg:consensus-training}, Stage 2).}

\rebuttal{\paragraph{Reward.}
Directly using self-consistency as a reward for on-policy reinforcement learning can lead to reward hacking~\citep{shafayat2025selftrain}, where the model converges to mutually consistent but incorrect predictions, such as empty segmentation masks. To mitigate this issue, we precompute a consensus target $\widehat y$ for each unlabeled input (Section~\ref{sec:multi-seed-voting}) and keep it fixed throughout training. These targets are generated by the SC-FT student, which therefore serves as the teacher for the subsequent SC-RL stage.

Following the early-readout procedure in Eq.~\ref{eq:early-consensus}, the current policy $F_{\theta_{\mathrm{old}}}$ generates $G$ rollouts $\{\widehat z_{\mathrm{clean},k}^i\}_{i=1}^{G}$ per input. Task-specific predictions are extracted at denoising step $k$:

\begin{equation}
    y_k^i = L\!\left(D(\widehat z_{\mathrm{clean},k}^i),c\right),
    \quad 1 \leq i \leq G.
    \label{eq:sc-consensus}
\end{equation}

Each rollout is then rewarded according to its agreement with the fixed consensus target:

\begin{equation}
    R_i^{\mathrm{SC}} = S(y_k^i,\widehat y).
    \label{eq:sc_reward_art}
\end{equation}

The similarity function $S$ is task-specific: exact path matching for maze solving, spatial agreement within radius $r$ of the consensus point (or a correct abstention when no target is identified) for visual search, and mask overlap for segmentation.}

\rebuttal{\paragraph{Policy update.}

To efficiently optimize the self-consistency reward in Eq.~\ref{eq:sc_reward_art}, we adopt DiffusionART~\citep{diffusionart} instead of GRPO-style policy optimization~\citep{shao2024deepseekmath,liu2025flow}. DiffusionART constructs a reward-guided update direction in latent space from the covariance between rollout rewards and predicted clean latents. Specifically, the direction is computed as
\begin{equation}
\begin{gathered}
\mathbf{C}(c) = \frac{1}{G}\sum_{i=1}^{G}
\big(R_i^{\mathrm{SC}}-\overline{R}^{\mathrm{SC}}\big)
\big(\widehat z^{\,i}-\overline{z}\big),\\
\mathbf{G}_{\mathrm{SC}}(c) =
\frac{\mathbf{C}(c)}
{\operatorname{RMS}\big(\mathbf{C}(c)\big)}.
\end{gathered}
\label{eq:sc_art_guidance}
\end{equation}
where $\overline{R}^{\mathrm{SC}}$ and $\overline z$ denote the mean reward and predicted clean latent across the $G$ rollouts, respectively.

At sampled denoising states $s_t=(z_t,t,c)$, we train the adapter to match a reward-shifted velocity target,
$v_{\mathrm{tgt}}(s_t)=v_{\theta_{\mathrm{old}}}(s_t)+\mathbf{G}_{\mathrm{SC}}(c)$
(Appendix~\ref{app:training}); $\mathbf{G}_{\mathrm{SC}}$ has unit RMS, so no step-size coefficient is needed.

Importantly, this optimization procedure is agnostic to the reward source. Replacing the self-consistency reward with a ground-truth verifier yields reinforcement learning with verifiable rewards (RLVR)~\citep{zhu2026video,liu2026wanr1}, which we evaluate in an ablation study in Section~\ref{sec:ablations}.
}

\section{Experiments}
\label{sec:experiments}

We conduct extensive experiments on three video reasoning tasks: referring expression segmentation, maze solving, and visual search.
Section~\ref{sec:experimental-setup} details the experimental setup.
Sections~\ref{sec:inference-results} and~\ref{sec:posttraining-results} report self-consistency results for inference and post-training, respectively.
Section~\ref{sec:qualitative-results} shows qualitative results, and Section~\ref{sec:ablations} presents an ablation study and discusses limitations and future work.

\begin{table*}[t!]
\centering
\color{revorange}
\caption{Inference with ten rollouts of the frozen MiniMax-H3 model on 300 search arrays (75 per condition), 270 validation maze layouts (90 per size), and 291 segmentation images; every row uses the same inputs.
Search and maze scores are percentages; segmentation scores are fractions.
\emph{Self-consistency} is the spatial vote for search, the modal path for mazes, and the pixel vote for segmentation.
\emph{VLM best-of-10} selects among the rollouts with an external VLM judge(Qwen3-VL-8B).
$\uparrow$: higher is better. Bold: ours. Notes on individual rows are in Appendix~\ref{app:table-notes}.}
\label{tab:inference-main}
\small
\setlength{\tabcolsep}{6.5pt}
\begin{tabular}{@{}lccccccccc@{}}
\toprule
 & \multicolumn{3}{c}{Visual search} & \multicolumn{3}{c}{Maze solving} & \multicolumn{3}{c}{Referring segmentation} \\
\cmidrule(lr){2-4}\cmidrule(lr){5-7}\cmidrule(l){8-10}
Method & Acc.$\uparrow$ & Hit$\uparrow$ & Spec.$\uparrow$ & Strict$\uparrow$ & Goal$\uparrow$ & Short.$\uparrow$ & gIoU$\uparrow$ & cIoU$\uparrow$ & Prec.$\uparrow$ \\
\midrule
Single generation & 48.4 & 96.2 & 0.6 & 59.1 & 92.4 & 56.8 & 0.402 & 0.253 & 0.454 \\
VLM best-of-10 & 50.3 & 96.0 & 4.7 & 53.3 & 84.4 & 49.3 & 0.322 & 0.197 & 0.357 \\
Hidden-state ensemble~\cite{wang2026demystifying} & 51.0 & 98.0 & 4.0 & 51.1 & 82.6 & 51.1 & 0.417 & 0.283 & 0.470 \\
\textbf{Self-consistency (ours)} & \textbf{98.7} & \textbf{98.0} & \textbf{99.3} & \textbf{82.2} & \textbf{95.6} & \textbf{82.2} & \textbf{0.497} & \textbf{0.403} & \textbf{0.598} \\
\midrule
\multicolumn{10}{@{}l}{\textit{Reference models without video generation}} \\
Qwen3-VL-8B~\cite{bai2025qwen3vl} & 80.7 & 75.3 & 86.0 & 24.8 & 98.1 & 24.8 & 0.538 & 0.493 & 0.772 \\
LISA-13B~\cite{lai2024lisa} & -- & -- & -- & -- & -- & -- & 0.561 & 0.521 & 0.708 \\
GLaMM-RefSeg~\cite{rasheed2024glamm} & -- & -- & -- & -- & -- & -- & 0.519 & 0.474 & 0.711 \\
\bottomrule
\end{tabular}
\end{table*}

\subsection{Experimental Setup}
\label{sec:experimental-setup}

We use the first-frame- and text-conditioned FL2VA checkpoint of MiniMax-H3~\citep{minimax2026h3}, a 33B rectified-flow video transformer with 50 blocks, as our backbone.
All student models are post-trained using rank-16 LoRA adapters.
H3 full generation follows the native denoising process with $N=49$ steps.
Early readout and student evaluation decode the predicted clean latent after $k=20$ denoising steps, which substantially reduces inference cost while closely preserving generation quality.

\noindent\textbf{Tasks.} Frozen Lake~\citep{towers2024gymnasium}, used to evaluate video-model maze reasoning by \citet{newman2026videomodelsreasonearly}, tests whether a generated route connects a start and goal while avoiding holes.
Visual search tests target localization and abstention across four conditions combining 2D/3D stimuli with conjunctive/disjunctive feature rules~\citep{campbell2024binding}.
Referring expression segmentation uses positive gRefCOCO expressions~\citep{grefcoco} and covers single-instance and multi-instance queries.

Sampling, optimization, and consensus settings are in Appendix~\ref{app:inference} and~\ref{app:training}.
We report strict validity for mazes, task accuracy for search, and mean per-image IoU (gIoU) for segmentation.
Strict validity requires reaching the goal without entering a hole or making a non-adjacent jump.
Search accuracy requires localization within 25 input pixels when a target is present and no detected marker when it is absent; specificity measures the latter rate.
Segmentation cIoU pools intersection and union pixels, while precision, recall, and foreground area characterize the predicted masks.
Single-generation scores average repeated draws over fixed seeds for each input.

\subsection{Quantitative Results of Inference with Self-Consistency}
\label{sec:inference-results}

Aggregating structured predictions improves frozen-model performance across search, maze solving, and referring segmentation (Table~\ref{tab:inference-main}).
For visual search, a 9-of-10 location vote raises task accuracy from $48.4\%$ to $98.7\%$, combining a $98.0\%$ hit rate with $99.3\%$ specificity.
The key benefit is reliable abstention: spurious markers vary across generations, allowing the vote to reject locations without recurring support.
For mazes, modal trajectory selection raises strict validity from $59.1\%$ to $82.2\%$.
The selected paths are shortest valid routes on $82.2\%$ of layouts.
For referring segmentation, 9-of-10 mask voting raises gIoU from $0.402$ to $0.497$ and precision from $0.454$ to $0.598$.
%
Table~\ref{tab:inference-main} also compares our consensus rules against external selection baselines on the same ten rollouts.
%
%
%
Additionally, hidden states ensemble average the hidden states of three seeds at the first denoising step~\cite{wang2026demystifying}, reaching $0.417$ gIoU on segmentation, below the pixel vote at $0.497$; it is close to a single generation on search ($51.0\%$ accuracy) and below it on mazes ($51.1\%$ strict validity).
%

\begin{table*}[t!]
\centering
\color{revorange}
\caption{Single-generation performance after post-training, with 20 denoising steps, on held-out inputs: 300 search arrays, 270 validation mazes,
and 291 images.
Search and maze scores are percentages; segmentation scores are fractions. The Base row is the single-generation row of Table~\ref{tab:inference-main}.
\emph{SC-FT}: self-consistency fine-tuning (Section~\ref{sec:self-consistency-posttraining}), which for search includes the counterfactual pairs of Appendix~\ref{app:training}; \emph{SC-RL}: self-consistency reinforcement learning (Section~\ref{sec:sc_art}).
$\uparrow$ / $\downarrow$: higher / lower is better; \emph{Area}, the predicted foreground fraction, has no preferred direction. Bold: best. Notes on individual entries are in Appendix~\ref{app:table-notes}.}
\label{tab:posttraining-main}
\small
\setlength{\tabcolsep}{7pt}
\begin{tabular}{@{}lcccccccccc@{}}
\toprule
 & \multicolumn{3}{c}{Visual search} & \multicolumn{4}{c}{Maze solving} & \multicolumn{3}{c}{Referring segmentation} \\
\cmidrule(lr){2-4}\cmidrule(lr){5-8}\cmidrule(l){9-11}
Method & Acc.$\uparrow$ & Hit$\uparrow$ & Spec.$\uparrow$ & Strict$\uparrow$ & Goal$\uparrow$ & Hole$\downarrow$ & Jump$\downarrow$ & gIoU$\uparrow$ & Prec.$\uparrow$ & Area \\
\midrule
Base & 48.4 & 96.2 & 0.6 & 59.1 & 92.4 & 30.2 & 11.9 & 0.402 & 0.454 & 0.373 \\
SC-FT & 78.2 & 99.0 & 57.5 & 66.1 & 98.3 & 25.6 & 9.9 & 0.473 & 0.534 & 0.288 \\
\quad + SC-RL & \textbf{99.1} & \textbf{99.3} & \textbf{98.8} & \textbf{95.6} & \textbf{99.8} & \textbf{4.4} & \textbf{0.2} & \textbf{0.582} & \textbf{0.725} & 0.237 \\
\bottomrule
\end{tabular}
\end{table*}

\label{sec:early-results}
Consensus also becomes available well before denoising finishes.
Early readout gives nearly identical consensus quality on all three tasks; we report segmentation as the representative case (Appendix~\ref{app:inference} and~\ref{app:qualitative}).
Reading each trajectory after 20 of its 49 denoiser calls lowers consensus gIoU by only $0.002$ and single-generation gIoU by $0.001$.
Inference time falls from $164$ to $78$ seconds per rollout.

\vspace{-5pt}
\subsection{Quantitative Results of Learning with Self-Consistency}
\label{sec:posttraining-results}

\rebuttal{SC-FT improves single-generation performance across all three tasks (Table~\ref{tab:posttraining-main}). However, its performance eventually plateaus. To overcome this limitation, we introduce SC-RL, which further improves performance through reinforcement learning guided by self-consistency rewards.
Both students run 20 denoising steps at inference and are evaluated on the same held-out inputs. }

\noindent\textbf{Visual search.}
\label{sec:search-posttraining-results}
Without counterfactual data, SC-FT raises single-generation search accuracy from $48.4\%$ to $69.4\%$ (Appendix~\ref{app:ablation-details}).
It achieves a $100.0\%$ hit rate and increases specificity from $0.6\%$ to $38.8\%$.

Counterfactual-augmented training strengthens the abstention signal further.
We recolor the object under a teacher consensus point to the distractor color and relabel the modified input with the same frozen ten-seed teacher.
These examples use the target-construction procedure in Section~\ref{sec:self-consistency-posttraining}, with teacher abstentions yielding unchanged-video targets.
This raises accuracy to $78.2\%$ and specificity to $57.5\%$, with a $99.0\%$ hit rate (Table~\ref{tab:posttraining-main}).
Appendix~\ref{app:training} describes the data construction and the matched contrast-loss control.

\revorange{SC-RL continues from the counterfactual student, with a reward for agreeing with that student's own 9-of-10 vote, and is trained on the same arrays used for SC-FT.
A single generation then reaches $99.1\%$ accuracy, with a $99.3\%$ hit rate and $98.8\%$ specificity, against $78.2\%$ accuracy and $57.5\%$ specificity for SC-FT (Table~\ref{tab:posttraining-main}).

}

\noindent\textbf{Maze solving.}
\label{sec:maze-posttraining-results}
\revorange{The held-out mazes cover $4\times4$, $5\times5$, and $6\times6$ layouts (Table~\ref{tab:posttraining-main}).
SC-FT raises strict validity from $59.1\%$ to $66.1\%$, and its validity falls with layout size: $75.0\%$, $68.3\%$, and $55.0\%$ on the $4\times4$, $5\times5$, and $6\times6$ layouts.
SC-RL starts from this student and is rewarded for reproducing the consensus path on the same training layouts.
It reaches $95.6\%$ strict validity: hole entries fall from $25.6\%$ to $4.4\%$, and non-adjacent jumps from $9.9\%$ to $0.2\%$.
The gain holds on every size: SC-RL reaches $91.7\%$, $100.0\%$, and $95.0\%$ on the three sizes.}

\noindent\textbf{Referring expression segmentation.}
\label{sec:asc-pl-results}
\rebuttal{The combination of SC-FT and SC-RL raises gIoU from $0.402$ to $0.582$ (Table~\ref{tab:posttraining-main}).
Training uses consensus-filtered examples from images disjoint from the test set (Appendix~\ref{app:training}).
The improvement reflects more precise foreground predictions: precision rises from $0.454$ to $0.725$, the predicted foreground fraction falls from $0.373$ to $0.237$, and recall is $0.684$.}

\begin{figure*}[!t]
\centering
\includegraphics[width=\linewidth]{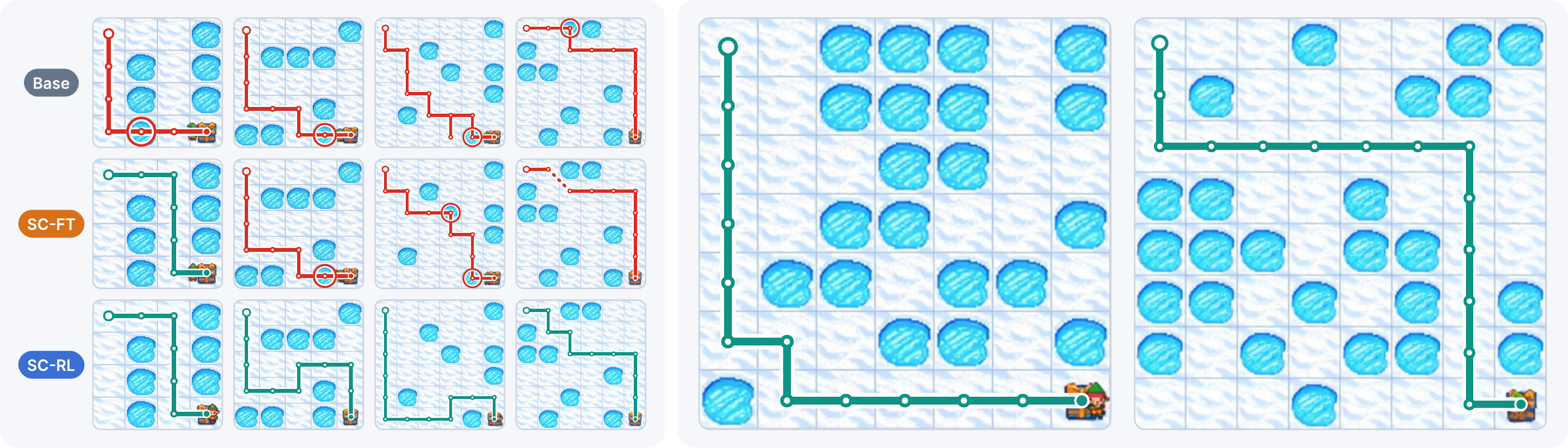}
\caption{Maze solving with a single generation.
Lines show the tracked path on the last frame (teal: valid; red: invalid; ring: entered hole; dots: jump).
(Left) Held-out layouts, same seed per column.
(Right) SC-RL on layout sizes absent from training.}
\label{fig:maze-qualitative}
\end{figure*}

\subsection{Qualitative Results}
\label{sec:qualitative-results}

\revorange{Figure~\ref{fig:maze-qualitative} (left) compares single generations on held-out mazes, with the same layout and seed for the three models in each column.
The frozen model steps into a hole in all four examples.
SC-FT repairs the $4\times4$ example and fails on the three larger ones: on the $5\times5$ layout it enters the same hole as the frozen model, on one $6\times6$ layout it enters two holes, and on the other it avoids the hole only by a diagonal jump.
SC-RL finds a valid route every time.
On the $5\times5$ layout it follows the frozen model's route for six cells, then turns away before the hole and reaches the goal two cells later; on the last layout it replaces the diagonal jump of SC-FT with two legal moves.
It also solves $7\times7$ and $8\times8$ layouts, sizes absent from its training (right): the two examples contain 21 and 25 holes, and both routes are shortest paths.
On these larger sizes, $81\%$ of its generations are strictly valid.}

\begin{figure}[t]
\centering
\includegraphics[width=\linewidth]{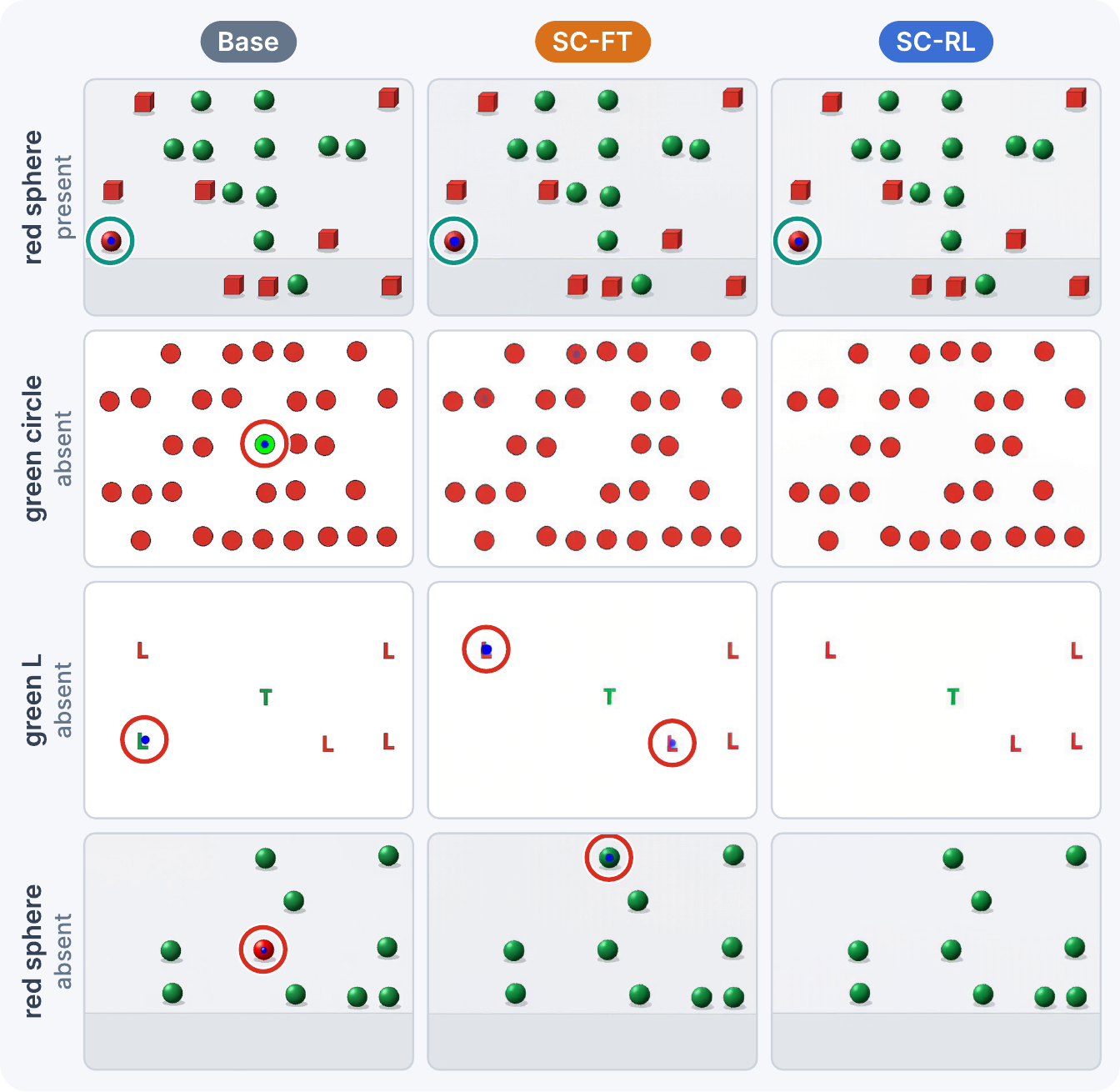}
\caption{Visual search with a single generation on held-out arrays.
Row labels give the target and whether it is present; rings show the generated markers (teal: on the target; red: no target in the input).}
\label{fig:search-qualitative}
\end{figure}

For referring segmentation, Figure~\ref{fig:teaser} (left) shows the typical behavior of the frozen model on the prompt ``luggage in front of person in brown sweater'': some rollouts paint only the luggage, and others also paint the people next to it.
The spill differs from seed to seed, so the pixel vote removes it and keeps the luggage.
After post-training, a single generation follows the referred region without the spill.
For visual search, Figure~\ref{fig:search-qualitative} shows one generation per model on four held-out arrays.
All three models mark the target when it is present.
When it is absent, the frozen model creates one, by adding an object or recoloring a distractor, and marks it.
SC-FT leaves the second array unchanged but still marks distractors in the last two, whereas SC-RL leaves all three unchanged.
Early readout leaves these outputs visually unchanged: masks decoded after 20 steps are nearly identical to those of the full 49-step generation.
Appendix~\ref{app:qualitative} shows more examples for all three tasks.

\subsection{Ablation Study}
\label{sec:ablations}

\begin{table*}[t!]
\centering
\caption{Ablations of the aggregation rule (top, ten rollouts of the frozen model) and of the fine-tuning targets (bottom, single generation).
\emph{Exact-match vote} counts identical predictions.
\emph{Random FT} trains on randomly chosen rollouts and \emph{Supervised FT} on rollouts selected with ground truth.
Panel (c) uses a separate set of $4\times4$ layouts, not the $4\times4$ mazes of Table~\ref{tab:posttraining-main}; Base in (d) is the single-generation row of (b).
Bold: ours. Appendix~\ref{app:ablation-details} lists further rows.}
\label{tab:ablation-main}
\footnotesize
\setlength{\tabcolsep}{3.5pt}
\begin{tabular*}{0.48\linewidth}[t]{@{\extracolsep{\fill}}lccc@{}}
\toprule
\multicolumn{4}{@{}l@{}}{\textit{(a) Search aggregation}, 300 arrays (\%)} \\
Method & Acc.$\uparrow$ & Hit$\uparrow$ & Spec.$\uparrow$ \\
\midrule
Single generation & 48.4 & 96.2 & 0.6 \\
Coordinate averaging & 42.0 & 84.0 & 0.0 \\
Exact-match vote & 76.0 & 90.0 & 62.0 \\
\textbf{Spatial vote} & \textbf{98.7} & \textbf{98.0} & \textbf{99.3} \\
\bottomrule
\end{tabular*}\hfill
\begin{tabular*}{0.48\linewidth}[t]{@{\extracolsep{\fill}}lccc@{}}
\toprule
\multicolumn{4}{@{}l@{}}{\textit{(b) Segmentation aggregation}, 291 images} \\
Method & gIoU$\uparrow$ & cIoU$\uparrow$ & Prec.$\uparrow$ \\
\midrule
Single generation & 0.402 & 0.253 & 0.454 \\
Exact-match vote & 0.012 & 0.033 & 0.013 \\
Most-consistent sample & 0.445 & 0.294 & 0.497 \\
\textbf{Pixel vote} & \textbf{0.497} & \textbf{0.403} & \textbf{0.598} \\
\bottomrule
\end{tabular*}

\vspace{6pt}
\begin{tabular*}{0.48\linewidth}[t]{@{\extracolsep{\fill}}lcccc@{}}
\toprule
\multicolumn{5}{@{}l@{}}{\textit{(c) Maze fine-tuning}, $4\times4$ layouts (\%)} \\
Method & Strict$\uparrow$ & Goal$\uparrow$ & Hole$\downarrow$ & Jump$\downarrow$ \\
\midrule
Base & 72.0 & 90.0 & 12.7 & 10.0 \\
Random FT & 72.7 & 84.7 & 8.0 & 10.0 \\
Supervised FT & 84.7 & 94.0 & 5.3 & 6.0 \\
\textbf{SC-FT} & \textbf{84.0} & \textbf{94.0} & \textbf{6.7} & \textbf{5.3} \\
\bottomrule
\end{tabular*}\hfill
\begin{tabular*}{0.48\linewidth}[t]{@{\extracolsep{\fill}}lcc@{}}
\toprule
\multicolumn{3}{@{}l@{}}{\textit{(d) Segmentation fine-tuning targets}, 291 images} \\
Training target & gIoU$\uparrow$ & Prec.$\uparrow$ \\
\midrule
Base & 0.402 & 0.454 \\
Most-consistent sample & 0.386 & 0.425 \\
Fixed sample (seed 0) & 0.441 & 0.484 \\
\textbf{Pixel consensus} & \textbf{0.473} & \textbf{0.534} \\
\bottomrule
\end{tabular*}
\end{table*}

\paragraph{Aggregation rule.}
\leavevmode\revorange{Table~\ref{tab:ablation-main}(a,b) compares the consensus rules with simpler aggregation on the same ten rollouts.
For search, averaging the marker coordinates fails ($42.0\%$) and exact-match voting reaches $76.0\%$, against $98.7\%$ for the spatial vote.
For segmentation, identical masks almost never occur ($0.012$ gIoU for exact match), and the single mask most consistent with the other nine reaches $0.445$ gIoU, below the pixel vote at $0.497$.
The support threshold and the number of rollouts are varied in Appendix~\ref{app:ablation-details}.}

\paragraph{Complementarity of SC-FT and SC-RL.}
Starting from the pretrained model, SC-FT and SC-RL independently improve gIoU from 40.2\% to 47.3\% and 53.1\%, respectively. Combining SC-FT with subsequent SC-RL further increases gIoU to 58.2\%, outperforming either method alone. These results demonstrate the complementary benefits of consensus-based fine-tuning and on-policy reward optimization.

\paragraph{Training targets.}
Table~\ref{tab:ablation-main}(c,d) asks whether the gains come from the consensus or from fine-tuning.
In the maze ablation, which uses a separate set of $4\times4$ layouts, SC-FT ($84.0\%$) exceeds fine-tuning on random rollouts ($72.7\%$) by $11.3$\% with the same layouts and rollout count, and is close to targets selected with ground truth ($84.7\%$); it recovers about $82\%$ of the gap between the base model's single generation ($72.0\%$) and its ten-seed modal vote ($86.7\%$).
For segmentation, training on the consensus mask ($0.473$ gIoU) outperforms training on a fixed sample ($0.441$) or on the most-consistent sample ($0.386$).
For search, counterfactual target-absent pairs raise accuracy from $69.4\%$ to $78.2\%$ (Appendix~\ref{app:ablation-details}).

\paragraph{Agreement versus correctness.}
We measure how well agreement tracks correctness.
It is reliable for search and mazes: a location marked by at least nine of the ten rollouts is almost always the target, and a path proposed by at least four rollouts is valid in $90.0\%$ of cases, against $24.8\%$ for a path proposed by one.
For segmentation it is weaker: a pixel marked by all ten rollouts lies inside the referent in only $64.6\%$ of cases (Appendix~\ref{app:ablation-details}).
As a reward, the consensus comes close to ground truth: under identical training settings, a ground-truth reward reaches $98.9\%$ strict validity on the maze layouts of Table~\ref{tab:posttraining-main}, where the consensus reward reaches $95.6\%$
(Appendix~\ref{app:training}).

\paragraph{Computational overhead.}
Training cost and inference time are reported in Appendix~\ref{app:training} and~\ref{app:inference}, respectively.

\subsection{Limitations and future work.}
\leavevmode\rebuttal{Self-consistency cannot identify systematic errors shared across rollouts, a limitation particularly evident in segmentation (Section~\ref{sec:ablations}). Extending the ground-truth-reward comparison to segmentation and investigating multiple valid solutions for other tasks remain directions for future work.}
Another direction is extension to physics-aware generation, for example a consensus-based reinforcement learning method that rewards rollouts for agreeing on the physical outcome.

\section{Conclusion}
Video generation is an emerging paradigm for visual reasoning and other vision tasks.
We study self-consistency as an intrinsic signal for improving diffusion-based video reasoning for both inference and post-training.
By aggregating structured predictions across stochastic rollouts, we show that self-consistency provides an effective training-free test-time scaling strategy for maze solving, visual search, and referring segmentation, while early readout substantially reduces its inference cost.
We further distill high-confidence consensus predictions into the video generator through Self-Consistency Fine-Tuning (SC-FT), enabling much of the multi-sample benefit to be recovered with a single rollout at test time.
Self-Consistency Reinforcement Learning (SC-RL) then rewards the student's own generations for agreeing with the consensus, which improves single-generation performance further.
Overall, our results show that agreement across stochastic generations can serve not only as an inference mechanism, but also as a scalable source of self-supervision for improving video-based perception and reasoning without ground-truth videos or task-specific verification.

\section*{LLM Usage Statement}
Generative AI tools, including OpenAI Codex and Google Gemini, assisted with writing polish and
technical support. All research concepts, ideas, and analyses were developed and conducted by the
authors. All AI-assisted outputs were manually reviewed and verified by the authors. Responsibility
for the final manuscript, implementation, experiments, and scientific claims rests with the authors.

{
    \small
    \bibliographystyle{ieeenat_fullname}
    \bibliography{main}
}

\clearpage
\appendix
\setcounter{figure}{0}\setcounter{table}{0}\setcounter{equation}{0}\setcounter{algorithm}{0}
\renewcommand{\thefigure}{S\arabic{figure}}
\renewcommand{\thetable}{S\arabic{table}}
\renewcommand{\theequation}{S\arabic{equation}}
\renewcommand{\thealgorithm}{S\arabic{algorithm}}
\captionsetup{font=small,labelfont=bf}
\renewcommand{\arraystretch}{1.06}
\setcounter{topnumber}{4}\setcounter{dbltopnumber}{4}\setcounter{totalnumber}{6}
\renewcommand{\topfraction}{0.9}
\renewcommand{\dbltopfraction}{0.9}
\renewcommand{\bottomfraction}{0.6}
\renewcommand{\textfraction}{0.08}
\renewcommand{\floatpagefraction}{0.8}
\renewcommand{\dblfloatpagefraction}{0.8}
\setcounter{topnumber}{3}
\setcounter{dbltopnumber}{3}
\setcounter{totalnumber}{5}
\makeatletter
\setlength{\@fptop}{0pt}
\setlength{\@dblfptop}{0pt}
\makeatother

\raggedbottom
\section{Inference Details}
\label{app:inference}

\begin{figure*}[t]
\centering
\setlength{\pw}{0.12\linewidth}
\setlength{\tabcolsep}{0pt}
\renewcommand{\arraystretch}{1}
\newcommand{\pvar}[1]{\textcolor{scft}{\textbf{#1}}}
\newcommand{\ptext}[1]{\parbox[c]{\dimexpr\linewidth-\pw-34pt\relax}{\sffamily\fontsize{9}{10.8}\selectfont\raggedright\vskip 6pt #1\par\vskip 6pt}}
\begin{tabular}{@{}c@{\hspace{5pt}}c@{\hspace{9pt}}l@{}}
\vlabel{\glabel{\textbf{Segmentation}}} & \gpanel{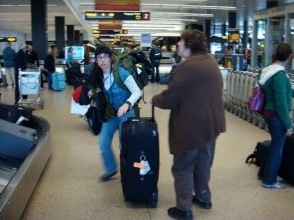} &
\ptext{Create a fixed-camera referring-expression segmentation video from the input photograph.\\[2pt]
Referring expression (verbatim): \pvar{``luggage in front of person in brown sweater''}\\[2pt]
Interpret the expression using the complete scene. Begin with the unchanged photograph, then gradually apply a precise semi-transparent red segmentation mask over every visible pixel belonging to all and only the object instance or object instances described by the expression. The expression may refer to one object, multiple objects, or no object. If nothing in the photograph matches it, keep the photograph unchanged and show no red mask.\\[2pt]
Preserve all original objects, people, geometry, lighting, and camera composition. Do not add, remove, move, recolor, or reshape anything. Finish with a stable, clean result.\\[2pt]
Strictly avoid camera movement, zoom, pan, crop, scene change, object motion, changed layout, added or removed objects, written text, captions, arrows, bounding boxes, masks on distractors, mask bleeding, missed target instances, blue mask, green mask, flicker, blur, low quality, and compression artifacts.} \\ \noalign{{\color{panelline}\hrule height 0.5pt}}
\vlabel{\glabel{\textbf{Search}}} & \gpanel{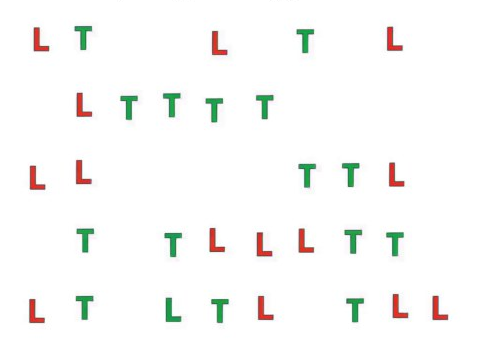} &
\ptext{Perform conjunctive visual search in the input image. The target is the \pvar{green letter L}. \pvar{Red L letters and green T letters} are distractors and must not be marked. Begin with the exact unchanged input image. If a \pvar{green L} is present, gradually add exactly one small, solid, bright blue dot at its center. If no \pvar{green L} is present, add no dot or other mark and keep the image unchanged. Preserve every original letter, color, position, and the white background. Static camera; no zoom, pan, crop, motion, text, arrows, outlines, new objects, removed objects, or changes other than the single requested blue dot. Finish with a stable frame. Silent audio.} \\ \noalign{{\color{panelline}\hrule height 0.5pt}}
\vlabel{\glabel{\textbf{Maze}}} & \gpanel{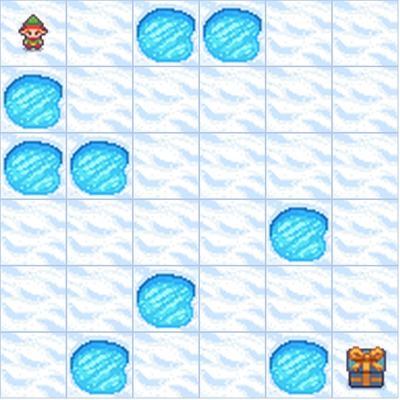} &
\ptext{Move the existing elf one orthogonally adjacent safe grid cell at a time until it touches the gift. Outside the dynamic region occupied by the elf, every pixel in every frame must remain exactly identical to the input image.} \\
\end{tabular}
\caption{\ctitle{Text prompts.}
One input image and its complete prompt for each task.
Orange text changes with the input: the referring expression for segmentation, and the target and distractors for search, where each of the four conditions has its own wording of the same instruction.
The maze prompt is the same for every layout.}
\label{fig:prompts}
\end{figure*}

\paragraph{Generation.}
All results use MiniMax-H3 FL2VA~\citep{minimax2026h3}, with the input image as the first frame, the task instruction as the text prompt (Figure~\ref{fig:prompts}), and ten seeds per input \meng{prompt for test-time scaling}.
Early readout stops at step $k=20$ of 49, decodes the predicted clean latent $\widehat z_{\mathrm{clean},k}^i=z_t^i+(1-t)v_\theta(z_t^i,t,c)$, and applies the same extractor and consensus rule as a full generation.
The model is never given ground-truth paths, target locations, or masks \meng{during inference}.

\begin{figure*}[t]
\centering
\includegraphics[width=\linewidth]{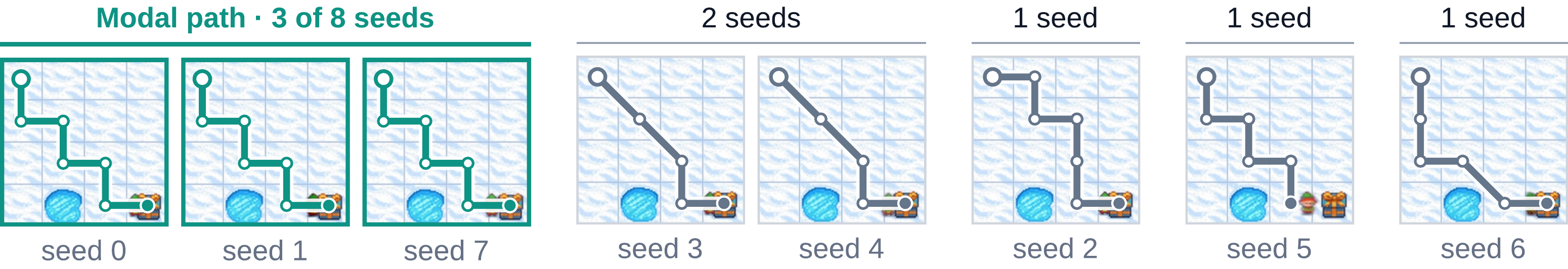}
\caption{\ctitle{Maze consensus on one training layout.}
The last frame of each of the eight sampled videos is overlaid with its extracted cell path.
The paths form five agreement groups, and the modal group (seeds 0, 1, and 7) is returned; its trajectories are reused as training targets for SC-FT.}
\label{fig:maze-consensus-example}
\end{figure*}

\paragraph{\meng{Result} extractors.}
Each task has one fixed extractor that turns a video into an answer.
For \emph{mazes}, SAM2.1~\citep{ravi2024sam2} tracks the agent from the first frame, each frame is assigned to the grid cell that holds most of the tracked mask, and repeated cells are merged into a cell path.
The modal rule returns the path that the most seeds produce (Figure~\ref{fig:maze-consensus-example}).
For \emph{visual search}, the model is asked to place a blue dot on the target and to leave the image unchanged when there is none; a fixed color filter detects the dot near the end of the video.
The vote groups dot locations within $r=25$ pixels and returns the median of the largest group when it has at least $m$ members, and no marker otherwise.
For \emph{referring segmentation}, the model paints the referred region with a red overlay.
Comparing the last frame with the first gives one binary mask per video, and the vote keeps the pixels that at least $m$ masks cover (Eq.~\ref{eq:pseudo_mask}).

\paragraph{Early readout.}
Table~\ref{tab:early-readout} compares early readout with full generation for segmentation.
Stopping at step 20 changes gIoU by at most $0.002$, for a single generation as well as for the consensus, and cuts the inference time per rollout from $164$ to $78$ seconds; Appendix~\ref{app:qualitative} shows the corresponding masks.
Early readout still decodes every rollout.
On a separate set of 100 expressions we also tried decoding one rollout only, the one whose early latent is closest to those of the other nine: it reaches $0.446$ gIoU, between a random rollout ($0.412$) and the vote over all ten masks ($0.483$).

\begin{table}[b]
\centering
\caption{\ctitle{Early readout versus full generation} on 100 segmentation expressions, with the same inputs and seeds.
\emph{Steps} is the number of denoising steps per rollout and \emph{Calls} the \meng{total} number of denoiser calls; time is in seconds on an NVIDIA A100.}
\label{tab:early-readout}
\footnotesize
\setlength{\tabcolsep}{2.5pt}
\begin{tabularx}{\linewidth}{lCCCCC}
\toprule
Method & Steps & Calls & gIoU$\uparrow$ & cIoU$\uparrow$ & Time$\downarrow$ \\
\midrule
Single generation & 49 & 49  & 0.373 & 0.224 & 164 \\
\ours \quad \meng{+} early readout & 20 & 20  & 0.372 & 0.224 & 78 \\
\midrule
Consensus & 49 & 490 & 0.489 & 0.359 & 1644 \\
\ours \quad \meng{+} early readout & 20 & 200 & 0.487 & 0.358 & 780 \\
\bottomrule
\end{tabularx}
\end{table}

\paragraph{Inference time.}
On an NVIDIA A100, MiniMax-H3 takes $164$\,s on average to generate one video with all 49 denoising steps and $78$\,s with early readout at step 20, excluding text embedding and reward computation; on an NVIDIA H100 the same measurement gives $92$\,s and $47$\,s.

\section{Post-Training Details}
\label{app:training}

\paragraph{\meng{Training Setup and Time}.}
All students use MiniMax-H3 FL2VA as the backbone, whose weights stay frozen. \meng{We} train rank-16 LoRA adapters~\citep{hu2022lora} on the attention and feed-forward projections of every transformer block with AdamW~\citep{loshchilov2019adamw}.
SC-FT uses learning rates between $2.5\times10^{-5}$ and $10^{-4}$ with four samples per update; SC-RL for search and mazes uses $3\times10^{-4}$ and samples eight rollouts per input in each update.
SC-FT requires around \textbf{350} H100 GPU hours for referring segmentation, and around \textbf{300} hours for visual search and maze solving.
For SC-RL to converge, referring segmentation requires around \textbf{400} H100 GPU hours, and visual search and maze solving around \textbf{200} each.

\paragraph{Consensus settings.}
Table~\ref{tab:consensus-configuration} lists the settings of the rules in Sections~\ref{sec:multi-seed-voting} and~\ref{sec:self-consistency-posttraining}.
Search and segmentation use $m=9$ of $G=10$ rollouts at inference and for the training targets; mazes use the path rule, with separate settings for the two.
$G$, $m$, and $r$ were chosen once on a small labeled development set and then kept fixed; neither the consensus nor the training reads ground truth.

\begin{table}[htbp]
\centering
\caption{\ctitle{Consensus settings.} $G$ is the number of rollouts that vote, $m$ the required support, and $r$ the grouping radius.}
\label{tab:consensus-configuration}
\footnotesize
\setlength{\tabcolsep}{4pt}
\begin{tabularx}{\linewidth}{@{}lccL@{}}
\toprule
Task & $G$ & $m$ & Rule \\
\midrule
Search & 10 & 9 & Spatial vote, $r=25$ pixels \\
Segmentation & 10 & 9 & Pixel vote \\
Mazes, inference & 10 & -- & Modal path \\
Mazes, training targets & 8 & 3 & Unique modal path \\
\bottomrule
\end{tabularx}
\end{table}

\begin{table*}[t]
\centering
\caption{\ctitle{Further ablation rows.}
(a) Aggregation rules on the ten rollouts of the frozen model.
(b, c) Fine-tuning, single generation with 20 denoising steps; in (c) all students share the inputs and the training recipe and differ only in the target, and the \emph{area-controlled sample} is the most-consistent sample shrunk to the area of the consensus mask.
Shaded: the settings used in the paper.}
\label{tab:ablation-full}
\footnotesize
\setlength{\tabcolsep}{3pt}
\begin{tabularx}{0.335\linewidth}[t]{lCCC}
\toprule
\multicolumn{4}{l}{\panelhead{(a) Search aggregation}, 300 arrays (\%)} \\
Method & Acc.$\uparrow$ & Hit$\uparrow$ & Spec.$\uparrow$ \\
\midrule
Single generation & 48.4 & 96.2 & 0.6 \\
Coordinate averaging & 42.0 & 84.0 & 0.0 \\
Exact-match vote & 76.0 & 90.0 & 62.0 \\
Spatial vote, 6 of 10 & 91.0 & 100.0 & 82.0 \\
\ours \textbf{Spatial vote, 9 of 10} & \textbf{98.7} & \textbf{98.0} & \textbf{99.3} \\
\bottomrule
\end{tabularx}\hfill
\begin{tabularx}{0.315\linewidth}[t]{lCCC}
\toprule
\multicolumn{4}{l}{\panelhead{(b) Search fine-tuning}, 300 arrays (\%)} \\
Method & Acc.$\uparrow$ & Hit$\uparrow$ & Spec.$\uparrow$ \\
\midrule
Base & 48.4 & 96.2 & 0.6 \\
SC-FT & 69.4 & 100.0 & 38.8 \\
\ours \quad \textbf{+ counterfactuals} & \textbf{78.2} & \textbf{99.0} & \textbf{57.5} \\
\bottomrule
\end{tabularx}\hfill
\begin{tabularx}{0.315\linewidth}[t]{lCC}
\toprule
\multicolumn{3}{l}{\panelhead{(c) Segmentation targets}, 291 images} \\
Training target & gIoU$\uparrow$ & Prec.$\uparrow$ \\
\midrule
Base & 0.402 & 0.454 \\
Most-consistent sample & 0.386 & 0.425 \\
Area-controlled sample & 0.423 & 0.471 \\
Fixed sample (seed 0) & 0.441 & 0.484 \\
\ours \textbf{Pixel consensus} & \textbf{0.473} & \textbf{0.534} \\
\bottomrule
\end{tabularx}
\end{table*}

\begin{table*}[t]
\centering
\caption{\ctitle{Notes on individual entries of the main tables.}}
\label{tab:table-notes}
\small
\renewcommand{\arraystretch}{1.12}
\begin{tabularx}{\linewidth}{@{}p{0.2\linewidth}X@{}}
\toprule
Entry & Note \\
\midrule
\rowcolor{oursrow}\multicolumn{2}{@{}l@{}}{\textbf{Table~\ref{tab:inference-main} (inference)}} \\
VLM best-of-10 & Qwen3-VL-8B picks one of the ten rollouts. For search the judge may also answer that the target is absent, in which case no marker is returned. \\
Qwen3-VL-8B & Answers directly from the input image. For segmentation it predicts a box, which SAM2.1-T converts to a mask. \\
LISA-13B, GLaMM-RefSeg & Specialist models trained with mask supervision on RefCOCO-family data. \\
\midrule
\rowcolor{oursrow}\multicolumn{2}{@{}l@{}}{\textbf{Table~\ref{tab:posttraining-main} (post-training)}} \\
SC-FT, SC-RL & Four generations per input. \\
SC-RL & Starts from the SC-FT student of the same task and is rewarded for agreeing with the fixed consensus on the training inputs of SC-FT. \\
\midrule
\rowcolor{oursrow}\multicolumn{2}{@{}l@{}}{\textbf{Table~\ref{tab:ablation-main} (ablations)}} \\
Teacher & On the test inputs, the ten-rollout consensus of the frozen model reaches $86.7\%$ strict validity on the layouts of panel (c) and $0.497$ gIoU on the images of panel (d). \\
\bottomrule
\end{tabularx}
\end{table*}

\paragraph{Training data.}
Training and evaluation use separate inputs in all three tasks.
For segmentation, the 9-of-10 pixel consensus of the frozen model on 211 training images gives the target masks, and each target video shows the input with the consensus mask fading in as a red overlay.
For search, the frozen model labels 80 training arrays, which cover all four conditions and ten object counts with and without a target; an array on which it abstains gives an unchanged-video target.
Part of the search training starts from the teacher's stored state $z_t$ instead of fresh noise, with the velocity target $u_t^\star=(z^\star-z_t)/(1-t)$.
For mazes, the frozen model generates eight videos for each new layout, and a layout is kept when its most frequent path is unique and produced by at least three of them; up to three of the matching trajectories become targets (Figure~\ref{fig:maze-consensus-example}).
The ablation of Table~\ref{tab:ablation-main}(c) uses $4\times4$ layouts only, with 60 held-out layouts for testing.

\begin{figure}[t]
\centering
\includegraphics[width=\linewidth]{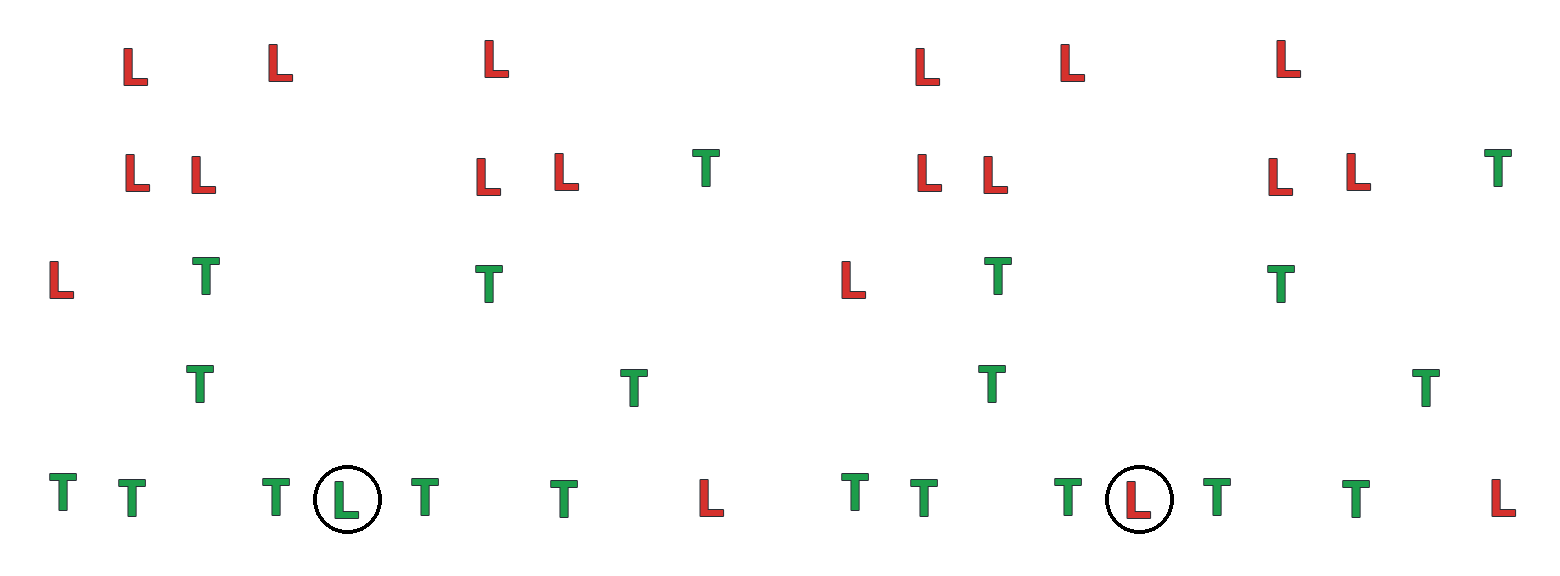}
\caption{\ctitle{A counterfactual training pair.}
The object at the teacher's consensus point is recolored (right), and the teacher abstains on the edited array, which gives an unchanged-video target.
The circle marks the edited object and is not part of the input.}
\label{fig:search-counterfactual}
\end{figure}

\paragraph{Counterfactual pairs for search.}
For each training array on which the teacher returns a point, we recolor the object at that point to the distractor color and leave everything else unchanged (Figure~\ref{fig:search-counterfactual}).
The frozen ten-seed teacher labels the edited array again, and the arrays on which it now abstains are added to the training pool.
No annotation is involved: the edit uses the teacher's own point, and the new label is the teacher's own vote.

\paragraph{Contrast-loss control.}
We also test whether the pairs need a dedicated loss: a pairwise term in the form of Diffusion-DPO~\citep{wallace2024diffusiondpo}, which prefers the teacher's target over a competing marker relative to a frozen reference adapter.
Adding it to the consensus loss on the same data brings no further gain in accuracy or specificity, so the counterfactual data alone accounts for the improvement.

\paragraph{SC-RL.}
SC-RL combines the self-consistency reward of Eq.~\ref{eq:sc_reward_art} with DiffusionART~\cite{diffusionart}.
The consensus $\widehat y$ is computed once before reinforcement learning and kept fixed, so the rollouts are scored and do not vote.
The adapter is trained to match the velocity target of Section~\ref{sec:sc_art} with a mean-squared error, using the timestep weighting and trust-region mask of DiffusionART.

\begin{figure*}[t]
\centering
\setlength{\pw}{\dimexpr(\linewidth-22pt)/8\relax}
\setlength{\tabcolsep}{1pt}
\renewcommand{\arraystretch}{1}
\newcommand{\quoteline}[1]{\multicolumn{4}{@{}p{\dimexpr4\pw+6pt\relax}@{\hspace{10pt}}}{\quotestyle ``#1''}}
\begin{tabular}{@{}cccc@{\hspace{10pt}}cccc@{}}
\multicolumn{2}{@{}c}{\glabel{Single generation}} & \multicolumn{2}{c@{\hspace{10pt}}}{\glabel{Consensus}} &
\multicolumn{2}{c}{\glabel{Single generation}} & \multicolumn{2}{c@{}}{\glabel{Consensus}} \\
\gnote{49 steps} & \gnote{20 steps} & \gnote{49 steps} & \gnote{20 steps} &
\gnote{49 steps} & \gnote{20 steps} & \gnote{49 steps} & \gnote{20 steps} \\[2pt]
\gpanel{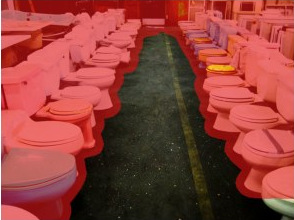} & \gpanel{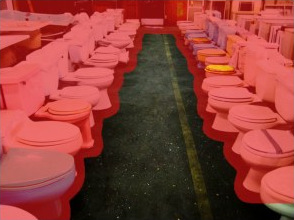} & \gpanel{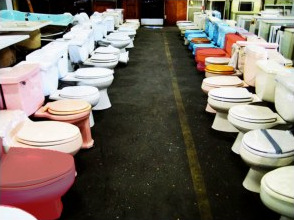} & \gpanel{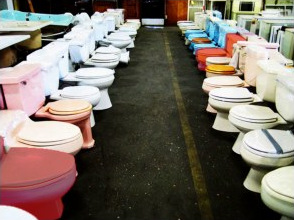} &
\gpanel{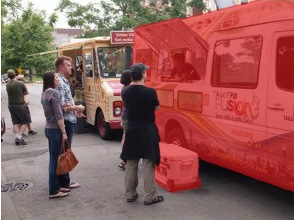} & \gpanel{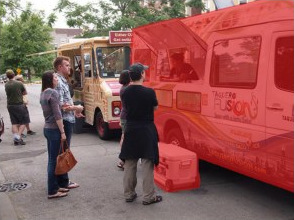} & \gpanel{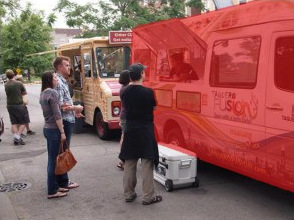} & \gpanel{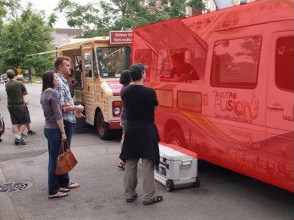} \\[1pt]
\gnote{IoU 0.11} & \gnote{0.11} & \glabel{\textcolor{scteal}{\textbf{0.89}}} & \glabel{\textcolor{scteal}{\textbf{0.89}}} &
\gnote{IoU 0.89} & \gnote{0.90} & \glabel{\textcolor{scteal}{\textbf{0.95}}} & \glabel{\textcolor{scteal}{\textbf{0.96}}} \\[-3pt]
\quoteline{a blue toilet bowl nearest to us on the left column and a yellow toilet bowl far away from us on the right column} &
\multicolumn{4}{@{}p{\dimexpr4\pw+6pt\relax}@{}}{\quotestyle ``the white bus in the right side near us and the brown bag carried by the lady standing in the left side''} \\[6pt]
\gpanel{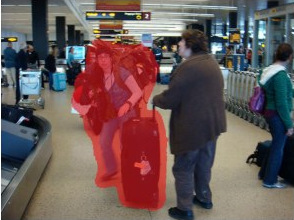} & \gpanel{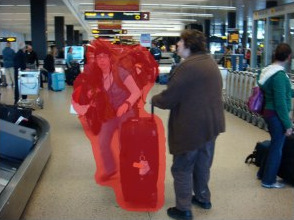} & \gpanel{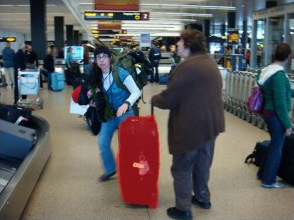} & \gpanel{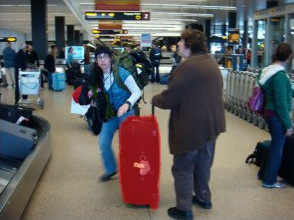} &
\gpanel{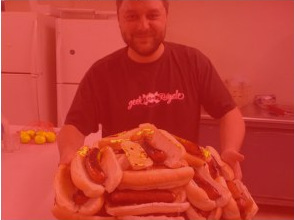} & \gpanel{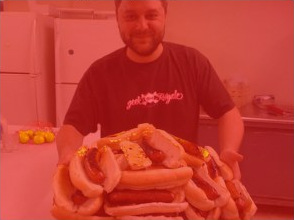} & \gpanel{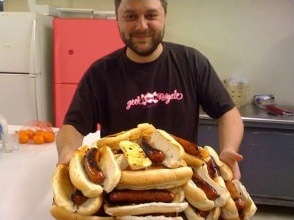} & \gpanel{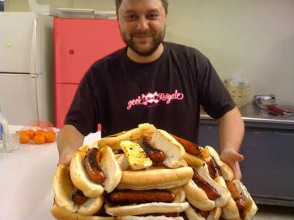} \\[1pt]
\gnote{IoU 0.28} & \gnote{0.28} & \glabel{\textcolor{scteal}{\textbf{0.88}}} & \glabel{\textcolor{scteal}{\textbf{0.88}}} &
\gnote{IoU 0.07} & \gnote{0.07} & \glabel{\textcolor{scteal}{\textbf{0.86}}} & \glabel{\textcolor{scteal}{\textbf{0.86}}} \\[-3pt]
\quoteline{luggage in front of person in brown sweater} &
\multicolumn{4}{@{}p{\dimexpr4\pw+6pt\relax}@{}}{\quotestyle ``the fridge behind the man on the right and the oranges in the left side on the table behind''} \\
\end{tabular}
\caption{\ctitle{Consensus versus single generation, and full generation versus early readout.}
Each block of four panels is one expression: a single generation and the 9-of-10 vote over ten rollouts, each decoded after the full 49 denoising steps and after 20 steps.
Numbers are the IoU with the ground truth.}
\label{fig:early-readout-gallery}
\end{figure*}

\begin{figure*}[t]
\centering
\setlength{\pw}{\dimexpr(\linewidth-22pt)/8\relax}
\setlength{\tabcolsep}{1pt}
\renewcommand{\arraystretch}{1}
\newcommand{\segrow}[2]{\gpanel{figures/supp/seg/seg#1_input.jpg} & \gpanel{figures/supp/seg/seg#1_base.jpg} & \gpanel{figures/supp/seg/seg#1_scft.jpg} & \gpanel{figures/supp/seg/seg#1_gt.jpg} &
\gpanel{figures/supp/seg/seg#2_input.jpg} & \gpanel{figures/supp/seg/seg#2_base.jpg} & \gpanel{figures/supp/seg/seg#2_scft.jpg} & \gpanel{figures/supp/seg/seg#2_gt.jpg}}
\newcommand{\iourow}[4]{& \gnote{IoU #1} & \glabel{\textcolor{scft}{\textbf{IoU #2}}} & & & \gnote{IoU #3} & \glabel{\textcolor{scft}{\textbf{IoU #4}}} &}
\newcommand{\quotepair}[2]{\multicolumn{4}{@{}p{\dimexpr4\pw+6pt\relax}@{\hspace{10pt}}}{\quotestyle ``#1''} &
\multicolumn{4}{@{}p{\dimexpr4\pw+6pt\relax}@{}}{\quotestyle ``#2''}}
\begin{tabular}{@{}cccc@{\hspace{10pt}}cccc@{}}
\glabel{Input} & \pill{scbase}{Base} & \pill{scft}{SC-FT} & \glabel{Ground truth} &
\glabel{Input} & \pill{scbase}{Base} & \pill{scft}{SC-FT} & \glabel{Ground truth} \\[1pt]
\segrow{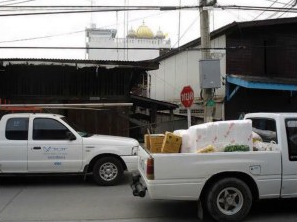}{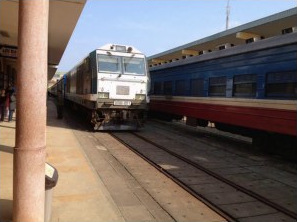} \\[1pt]
\iourow{0.21}{0.91}{0.16}{0.91} \\[-3pt]
\quotepair{the white truck in the front near us and the red sign in the background right}{white train} \\[6pt]
\segrow{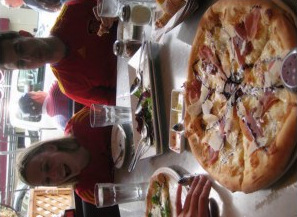}{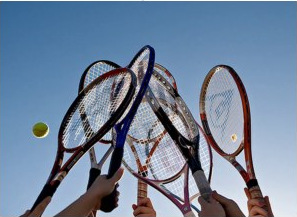} \\[1pt]
\iourow{0.22}{0.88}{0.17}{0.84} \\[-3pt]
\quotepair{bottom rightmost pizza slice as we view the pic}{the green tennis ball in the left side flying in the air and the racket on the right} \\[6pt]
\segrow{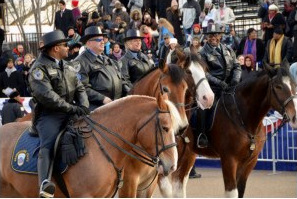}{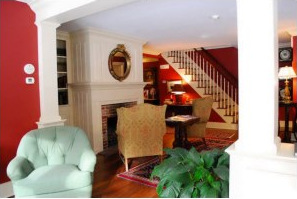} \\[1pt]
\iourow{0.17}{0.70}{0.22}{0.91} \\[-3pt]
\quotepair{horse on far right dark brown}{the green chair on the left bottom and the green plant on the right bottom corner} \\
\end{tabular}
\caption{\ctitle{Referring segmentation before and after SC-FT} on test images (seed 0, early readout).
Numbers are the IoU with the ground truth.}
\label{fig:segmentation-gallery}
\end{figure*}

\section{More Ablation Results}
\label{app:ablation-details}

Table~\ref{tab:ablation-full} lists the ablation rows that Table~\ref{tab:ablation-main} leaves out.


\paragraph{Aggregation and fine-tuning.}
For search, a strict majority (6 of 10) already reaches $91.0\%$ accuracy, and the 9-of-10 rule gains the rest by rejecting the remaining scattered markers (Table~\ref{tab:ablation-full}a); panel (b) lists SC-FT without the counterfactual pairs.
For segmentation, a student trained on a single sample that is shrunk to the area of the consensus mask stays at $0.423$ gIoU, so the gain of the consensus target does not come from smaller masks (Table~\ref{tab:ablation-full}c).

\begin{table}[!htb]
\centering
\caption{\ctitle{Segmentation gIoU for $G$ rollouts and support $m$} on the 100 expressions of Table~\ref{tab:early-readout}, with early readout.
Darker cells are higher; bold entries mark the rule $m=\lceil0.9G\rceil$.}
\label{tab:ablation-support}
\scriptsize
\setlength{\tabcolsep}{0pt}
\renewcommand{\arraystretch}{1.25}
\begin{tabularx}{\linewidth}{>{\centering\arraybackslash}p{0.085\linewidth}CCCCCCCCCC}
\toprule
$G\backslash m$ & 1 & 2 & 3 & 4 & 5 & 6 & 7 & 8 & 9 & 10 \\
\midrule
1 & \cellcolor{scteal!25}\textbf{.372} &  &  &  &  &  &  &  &  &  \\
2 & \cellcolor{scteal!17}.316 & \cellcolor{scteal!32}\textbf{.423} &  &  &  &  &  &  &  &  \\
3 & \cellcolor{scteal!12}.283 & \cellcolor{scteal!27}.383 & \cellcolor{scteal!35}\textbf{.443} &  &  &  &  &  &  &  \\
4 & \cellcolor{scteal!9}.262 & \cellcolor{scteal!21}.348 & \cellcolor{scteal!32}.418 & \cellcolor{scteal!36}\textbf{.452} &  &  &  &  &  &  \\
5 & \cellcolor{scteal!7}.247 & \cellcolor{scteal!18}.323 & \cellcolor{scteal!27}.384 & \cellcolor{scteal!35}.441 & \cellcolor{scteal!37}\textbf{.455} &  &  &  &  &  \\
6 & \cellcolor{scteal!5}.236 & \cellcolor{scteal!15}.306 & \cellcolor{scteal!23}.359 & \cellcolor{scteal!30}.409 & \cellcolor{scteal!37}.456 & \cellcolor{scteal!37}\textbf{.454} &  &  &  &  \\
7 & \cellcolor{scteal!4}.226 & \cellcolor{scteal!13}.293 & \cellcolor{scteal!20}.339 & \cellcolor{scteal!27}.384 & \cellcolor{scteal!33}.429 & \cellcolor{scteal!39}.467 & \cellcolor{scteal!36}\textbf{.452} &  &  &  \\
8 & \cellcolor{scteal!3}.218 & \cellcolor{scteal!12}.284 & \cellcolor{scteal!18}.323 & \cellcolor{scteal!24}.363 & \cellcolor{scteal!30}.405 & \cellcolor{scteal!35}.444 & \cellcolor{scteal!40}.476 & \cellcolor{scteal!36}\textbf{.449} &  &  \\
9 & \cellcolor{scteal!2}.211 & \cellcolor{scteal!11}.276 & \cellcolor{scteal!16}.312 & \cellcolor{scteal!21}.346 & \cellcolor{scteal!27}.384 & \cellcolor{scteal!32}.423 & \cellcolor{scteal!37}.454 & \cellcolor{scteal!41}.482 & \cellcolor{scteal!35}\textbf{.445} &  \\
10 & \cellcolor{scteal!1}.205 & \cellcolor{scteal!10}.269 & \cellcolor{scteal!15}.304 & \cellcolor{scteal!19}.330 & \cellcolor{scteal!24}.366 & \cellcolor{scteal!29}.402 & \cellcolor{scteal!34}.436 & \cellcolor{scteal!38}.462 & \cellcolor{scteal!42}\textbf{.487} & \cellcolor{scteal!35}.440 \\
\bottomrule
\end{tabularx}
\end{table}

\paragraph{Number of rollouts and support.}
Table~\ref{tab:ablation-support} varies the number of rollouts $G$ and the support $m$ for segmentation, averaging over every subset of size $G$ of the ten seeds.
For every $G$ a high threshold works best, and the best setting is the 9-of-10 rule used in the paper.
Requiring all ten lowers gIoU from $0.487$ to $0.440$, so the rule benefits from tolerating one dissenting rollout; up to nine rollouts $\lceil0.9G\rceil$ still asks for unanimity, which is why the value at ten exceeds the one at nine.

\paragraph{Agreement versus correctness.}
For search, no location marked by fewer than eight of the ten rollouts is the target.
For mazes, validity rises with support, from $24.8\%$ for a path proposed by a single rollout to $90.0\%$ for one proposed by at least four.
For segmentation, a pixel marked by all ten rollouts lies inside the referent in $64.6\%$ of cases, because the rollouts can agree on a region that is not the referent.

\paragraph{\meng{Comparison to} ground-truth reward.}
For mazes, \meng{we ablate} SC-RL with the same starting student, layouts, and settings, and reward a rollout when its path is strictly valid under the true layout instead of when it equals the consensus path.
It reaches $98.9\%$ strict validity on the validation layouts of Table~\ref{tab:posttraining-main}, against $95.6\%$ for the consensus reward.
\meng{Using our proposed consensus-based reward is almost as good as the ground-truth reward in this case.}
Appendix~\ref{app:qualitative} shows layouts on which the two differ. 
\section{Notes on the Main Tables}
\label{app:table-notes}

Table~\ref{tab:table-notes} gives notes on individual entries of Tables~\ref{tab:inference-main}, \ref{tab:posttraining-main}, and~\ref{tab:ablation-main}.


\section{More Qualitative Results}
\label{app:qualitative}


\subsection{Referring Segmentation}
\label{app:qualitative-segmentation}

\begin{figure}[b]
\centering
\setlength{\pw}{\dimexpr(\linewidth-4pt)/2\relax}
\setlength{\tabcolsep}{2pt}
\begin{tabular}{@{}cc@{}}
\gpanel{} & \gpanel{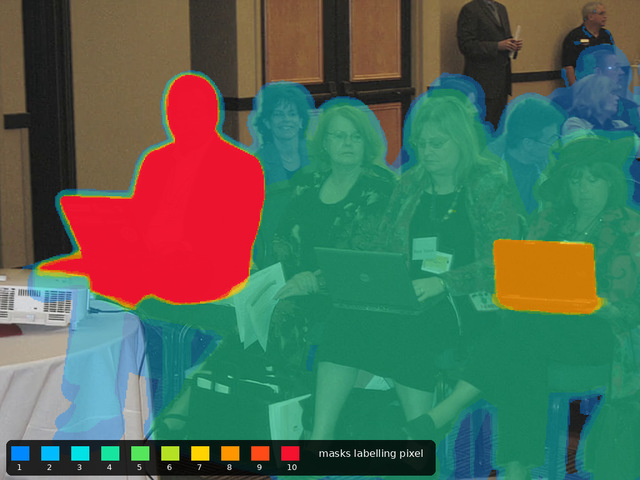} \\[2pt]
\glabel{Input} & \glabel{Votes over ten rollouts} \\
\end{tabular}
\caption{\ctitle{Pixel votes for one referring expression.}
Color shows how many of the ten extracted masks cover each pixel, from few (blue) to all ten (red).}
\label{fig:qualitative-outputs}
\end{figure}

The referred regions receive votes from almost every seed, while the spill-over changes from seed to seed and is removed by the 9-of-10 rule (Figures~\ref{fig:early-readout-gallery} and~\ref{fig:qualitative-outputs}).
The 20-step masks are nearly identical to the 49-step masks.
A single generation of the frozen model often paints most of the image or every object of the named category; after SC-FT it commits to the selected instance, and the errors that remain concern the extent of the mask more than its location (Figure~\ref{fig:segmentation-gallery}).

\subsection{Maze Solving}
\label{app:qualitative-maze}

\begin{figure}[!t]
\centering
\setlength{\pw}{\dimexpr(\linewidth-22pt)/4\relax}
\setlength{\tabcolsep}{1pt}
\renewcommand{\arraystretch}{1}
\newcommand{\routerow}[1]{\gpanel{figures/supp/maze/routes0_#1.jpg} & \gpanel{figures/supp/maze/routes1_#1.jpg} & \gpanel{figures/supp/maze/routes2_#1.jpg} & \gpanel{figures/supp/maze/routes3_#1.jpg}}
\begin{tabular}{@{}c@{\hspace{3pt}}cccc@{}}
\vlabel{\pill{scbase}{Base}} & \routerow{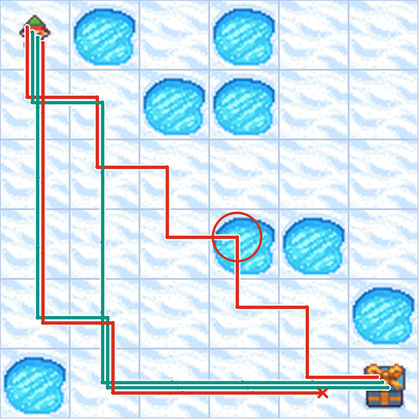} \\[2pt]
\vlabel{\pill{scft}{SC-FT}} & \routerow{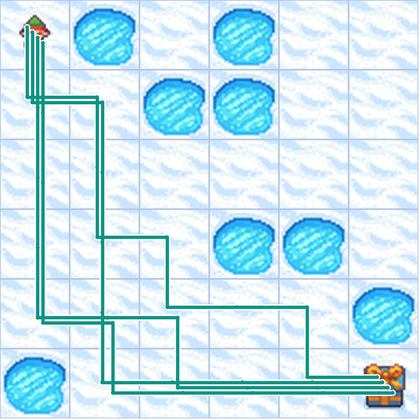} \\[2pt]
\vlabel{\pill{scrl}{SC-RL}} & \routerow{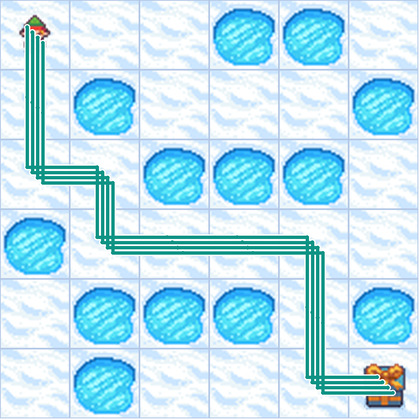} \\
\end{tabular}
\caption{\ctitle{All four generations of a layout, drawn together.}
Each column is one $6\times6$ validation layout; each line is the tracked path of one generation (teal: strictly valid; red: not; ring: entered hole; dotted: jump; cross: stopped before the goal).
Paths are shifted slightly so that repeated routes stay visible.}
\label{fig:maze-routes}
\end{figure}

\begin{figure}[!t]
\centering
\setlength{\pw}{\dimexpr(\linewidth-28pt)/3\relax}
\setlength{\tabcolsep}{1.5pt}
\renewcommand{\arraystretch}{1}
\begin{tabular}{@{}c@{\hspace{3pt}}ccc@{}}
\vlabel{\pill{scrl}{SC-RL}} & \gpanel{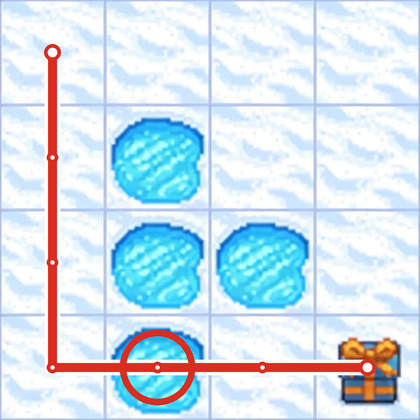} & \gpanel{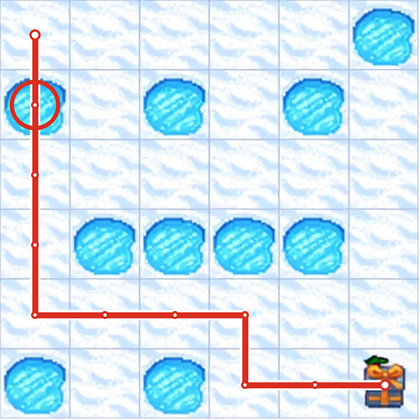} & \gpanel{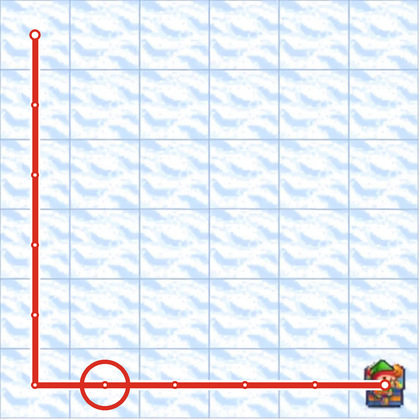} \\[2pt]
\vlabel{\pill{black!75}{Truth reward}} & \gpanel{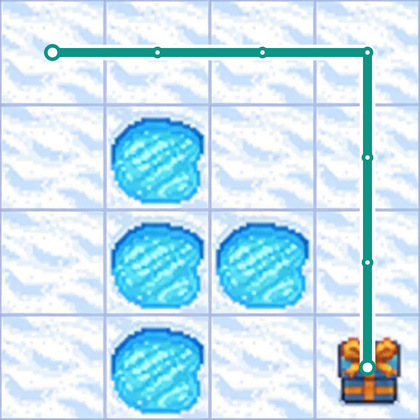} & \gpanel{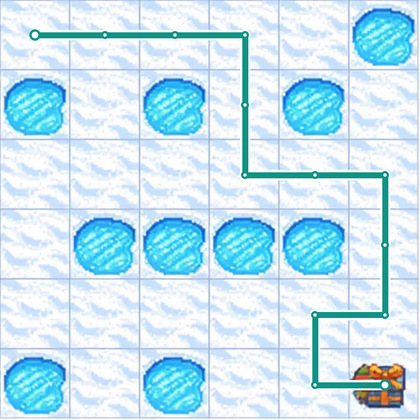} & \gpanel{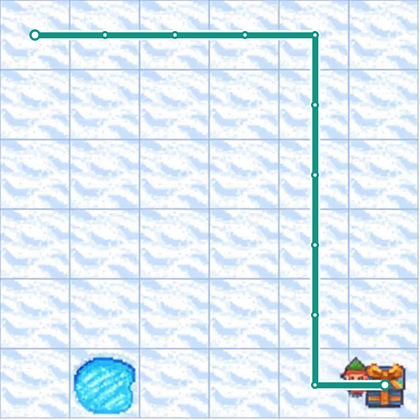} \\[2pt]
 & \glabel{(a)} & \glabel{(b)} & \glabel{(c)} \\
\end{tabular}
\caption{\ctitle{Failures that remain after SC-RL} (top) and the ground-truth-reward control on the same layout and rollout index (bottom).
(a) All four generations of this layout follow the route shown and enter the same hole.
(b) The path enters a hole on its first move.
(c) The path crosses the only hole of the layout, and the video then erases the hole.}
\label{fig:maze-failures}
\end{figure}

\begin{figure*}[t]
\centering
\setlength{\pw}{\dimexpr(\linewidth-132pt)/7\relax}
\setlength{\tabcolsep}{1pt}
\renewcommand{\arraystretch}{1}
\newcommand{\rowtag}[2]{\parbox[c]{50pt}{\raggedleft\sffamily\fontsize{9}{10.5}\selectfont\color{mutedtext}#1\\#2}}
\begin{tabular}{@{}r@{\hspace{4pt}}cccc@{\hspace{12pt}}r@{\hspace{4pt}}ccc@{}}
 & \multicolumn{4}{c}{\glabel{\textbf{(a) Fine-tuning}}} & & \multicolumn{3}{c}{\glabel{\textbf{(b) During SC-RL}}} \\[2pt]
 & \glabel{Input} & \pill{scbase}{Base} & \pill{scft}{SC-FT} & \pill{scft}{+ CF} & & \pill{scft}{0 updates} & \glabel{10 updates} & \pill{scrl}{20 updates} \\[2pt]
\rowtag{green L}{present} & \gpanel{figures/supp/search/gallery0_input.png} & \gpanelc{scred}{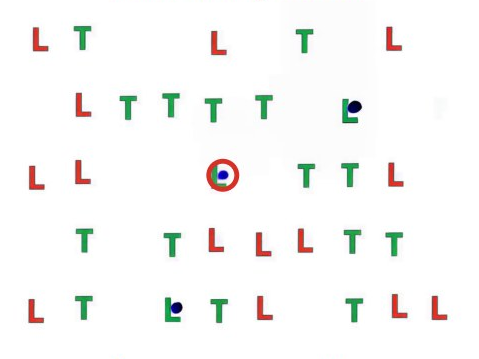} & \gpanelc{scteal}{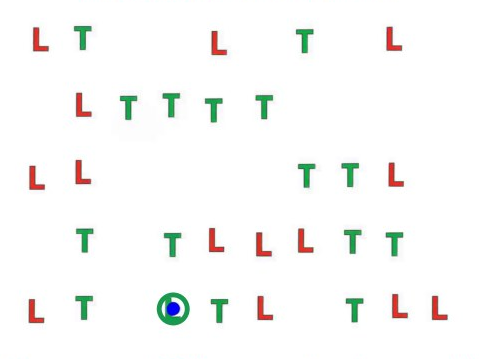} & \gpanelc{scteal}{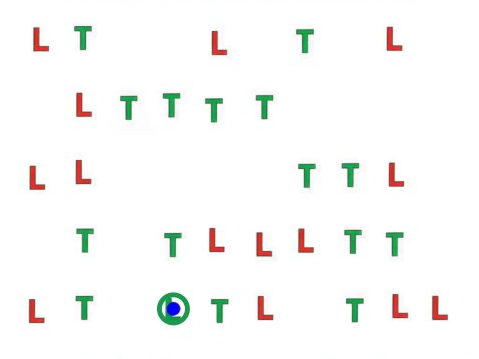} &
\rowtag{green L}{absent} & \gpanelc{scred}{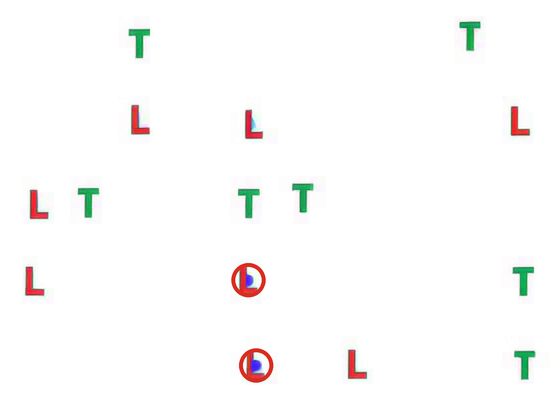} & \gpanelc{scred}{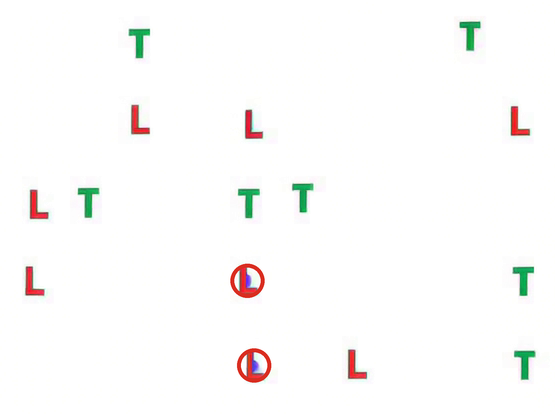} & \gpanelc{scteal}{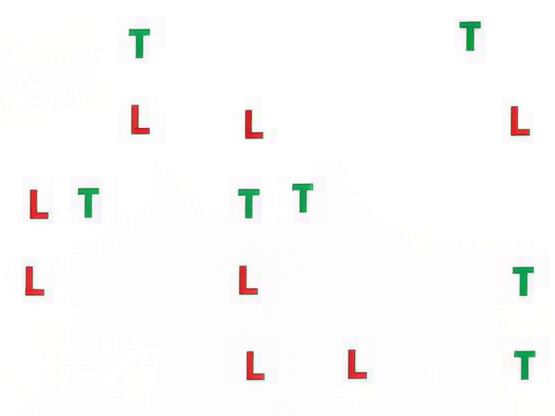} \\[2pt]
\rowtag{green circle}{absent} & \gpanel{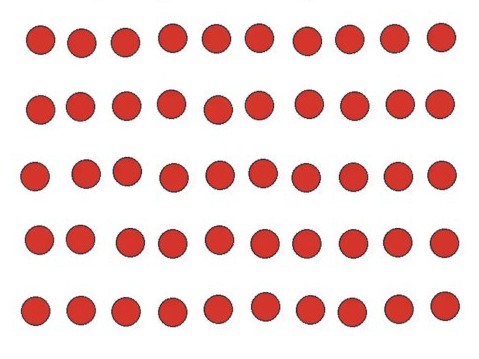} & \gpanelc{scred}{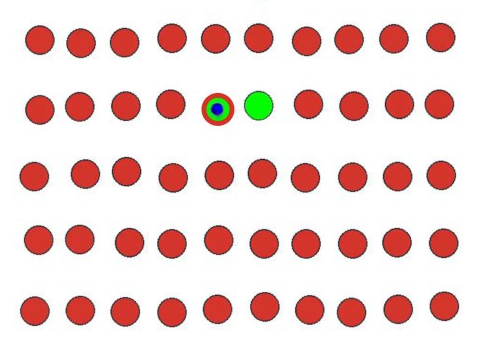} & \gpanelc{scteal}{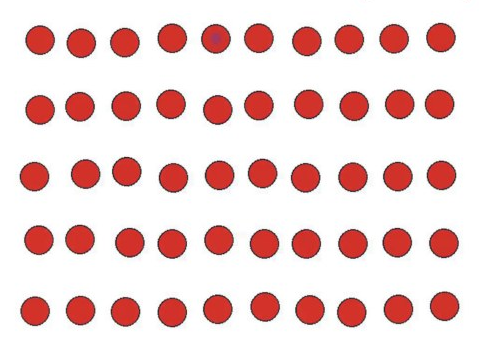} & \gpanelc{scteal}{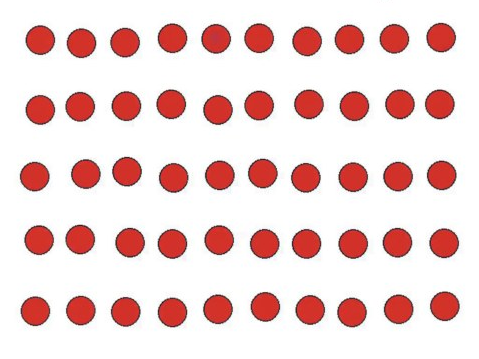} &
\rowtag{green circle}{absent} & \gpanelc{scred}{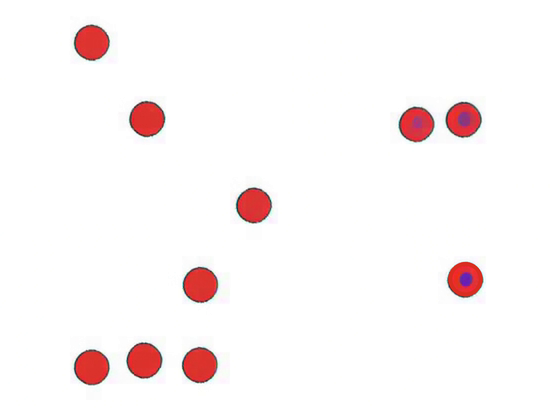} & \gpanelc{scteal}{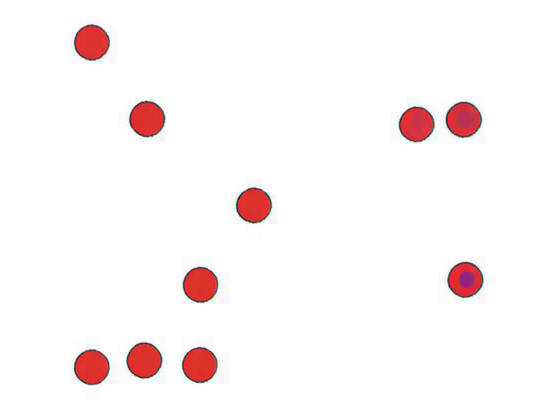} & \gpanelc{scteal}{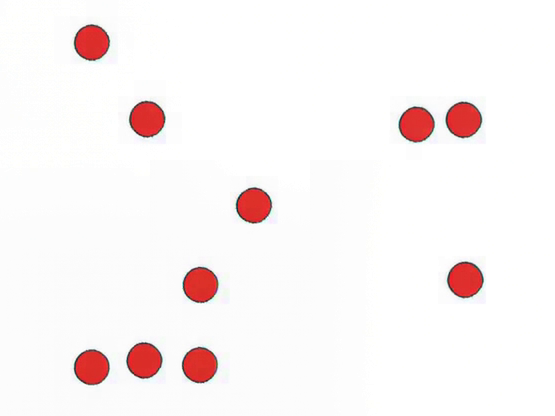} \\[2pt]
\rowtag{red sphere}{absent} & \gpanel{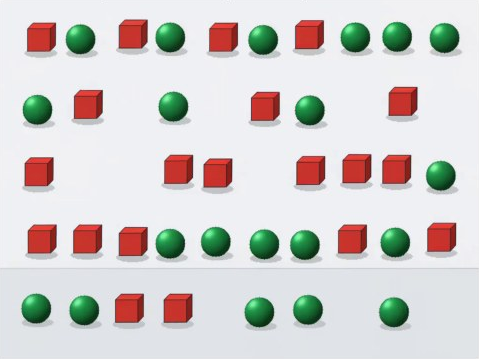} & \gpanelc{scred}{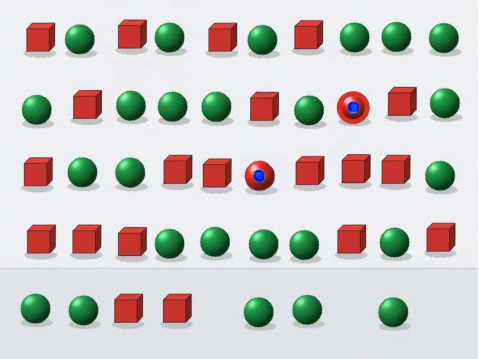} & \gpanelc{scteal}{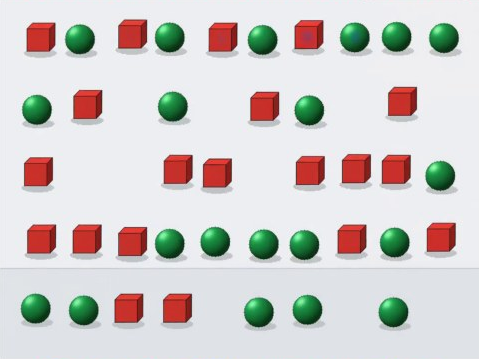} & \gpanelc{scteal}{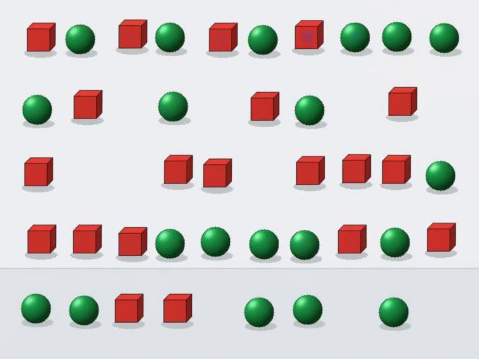} &
\rowtag{red sphere}{absent} & \gpanelc{scred}{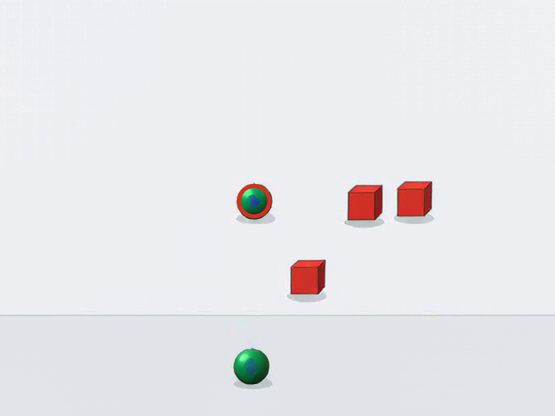} & \gpanelc{scteal}{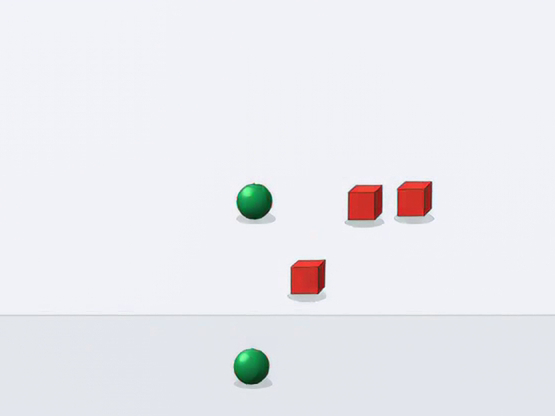} & \gpanelc{scteal}{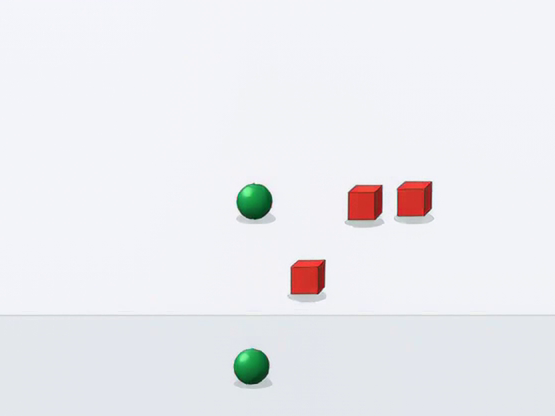} \\[2pt]
\rowtag{red sphere}{absent} & \gpanel{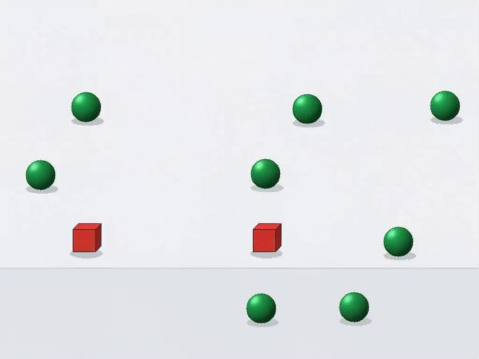} & \gpanelc{scred}{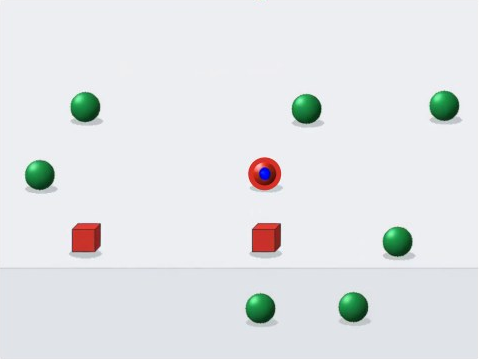} & \gpanelc{scred}{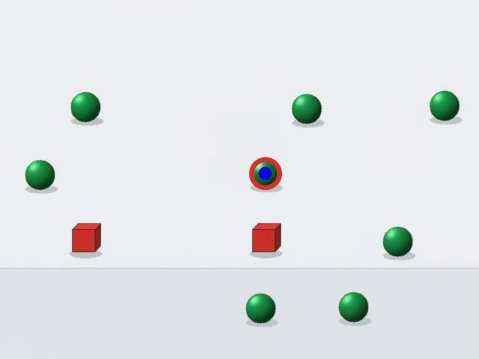} & \gpanelc{scteal}{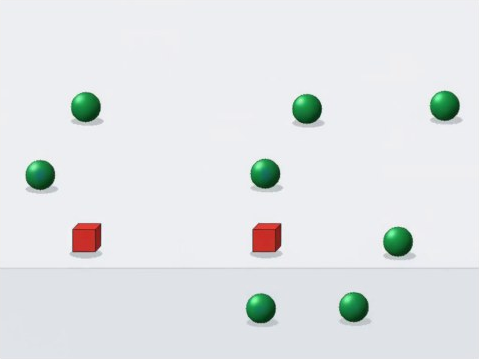} &
\rowtag{red sphere}{absent} & \gpanelc{scred}{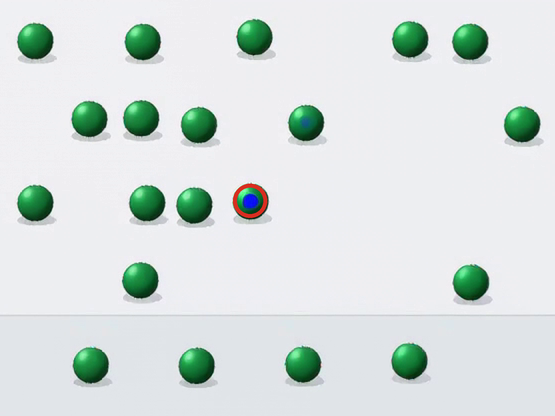} & \gpanelc{scred}{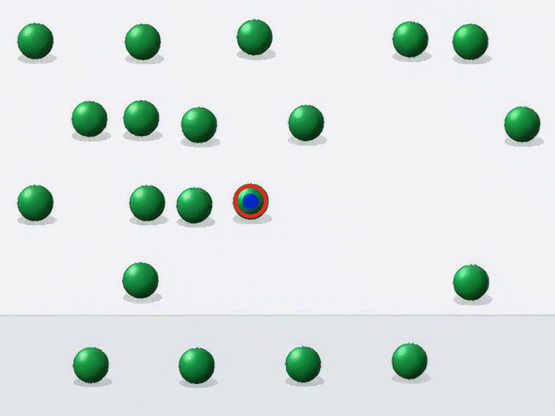} & \gpanelc{scteal}{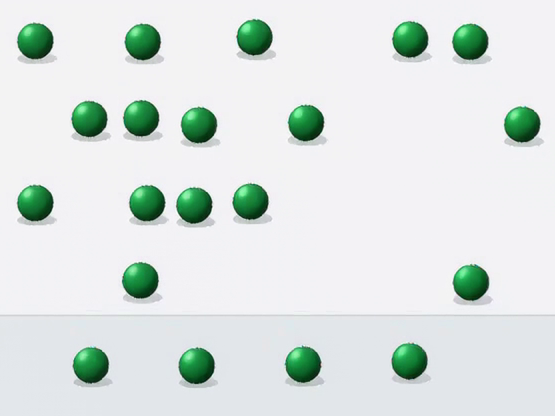} \\
\end{tabular}
\caption{\ctitle{Visual search with a single generation.}
Row labels give the target and whether it is present; rings mark the generated markers, and the frame of a panel is teal when the generation is correct and red when it is not.
(a) Test arrays before fine-tuning, after SC-FT, and after SC-FT with counterfactual pairs (+ CF), with the same seed in each row.
(b) Target-absent arrays during SC-RL, which starts from the student with counterfactual pairs (0 updates); each row shows the same rollout index after 0, 10, and 20 updates.}
\label{fig:search-gallery}
\end{figure*}

The four generations of the frozen model usually follow different routes, and several of them cross a hole; SC-FT makes more of them valid, and after SC-RL they mostly follow one and the same route (Figure~\ref{fig:maze-routes}).
The failures that are left are mostly hole entries shared by all four generations, which a reward for agreement cannot flag and which the ground-truth-reward control avoids (Figure~\ref{fig:maze-failures}).

\subsection{Visual Search}
\label{app:qualitative-search}

\begin{table}[!htb]
\centering
\caption{\ctitle{False markers on target-absent arrays during SC-RL.}
Number of generations that place a marker, out of 20 per condition (five held-out arrays, four generations each), after 0, 10, and 20 updates.
On the matching target-present arrays the marker is on the target in all 80 generations at each of the three points.}
\label{tab:search-false-markers}
\footnotesize
\begin{tabularx}{\linewidth}{lCCC}
\toprule
Condition & 0 updates & 10 updates & 20 updates \\
\midrule
2D conjunctive & 19 & 8 & 0 \\
2D disjunctive & 3 & 0 & 0 \\
3D conjunctive & 1 & 0 & 0 \\
3D disjunctive & 6 & 2 & 0 \\
\midrule
All (of 80) & 29 & 10 & 0 \\
\bottomrule
\end{tabularx}
\end{table}

On a target-absent array the frozen model usually produces the target first, by drawing a new object or recoloring a distractor, and then marks it (Figure~\ref{fig:search-gallery}a).
After SC-FT with counterfactual pairs, the errors that remain are small markers on target-absent arrays, most of them on 2D conjunctive arrays (Table~\ref{tab:search-false-markers}).
SC-RL removes them within twenty updates and leaves the markers on target-present arrays unaffected (Figure~\ref{fig:search-gallery}b).

Example videos generated with our methods are on the project page, \url{https://sc-video-reasoning.github.io}.

\FloatBarrier

\end{document}